\documentclass[10pt]{article}
\usepackage[top=0.75in, bottom=0.75in, left=0.75in, right=0.75in]{geometry}
\usepackage{palatino} 

\usepackage[numbers,sort&compress]{natbib}
\usepackage{setspace}
\usepackage{graphicx}
\usepackage{overpic,color}
\usepackage{float,subcaption,overpic}
\usepackage{algorithm}
\usepackage{amsmath,amsfonts,amscd,amssymb,graphicx,bm}
\usepackage{xfrac}
\usepackage{mathtools}
\usepackage{multirow,booktabs}
\usepackage{rotating}
\usepackage{subeqnar}
\usepackage[dvipsnames]{xcolor}

\graphicspath{{figures/}}
\newcommand{\todo}[1]{}
\newcommand{\ba}{\mathbf{a}}

\newcommand{\bPhi}{\mathbf{\Phi}}

\newcommand{\bphi}{\boldsymbol{\phi}}

\newcommand{\bC}{\mathbf{C}}
\newcommand{\bL}{\mathbf{L}}
\newcommand{\bN}{\mathbf{N}}
\newcommand{\bS}{\mathbf{S}}

\newcommand{\bV}{\mathbf{V}}

\newcommand{\bx}{\mathbf{x}}
\newcommand{\bX}{\mathbf{X}}

\newcommand{\by}{\mathbf{y}}

\newcommand{\reals}{\mathbb{R}}

\usepackage{etoolbox}
\usepackage{tikz}

\DeclareMathOperator*{\argmin}{arg\,min}

\usepackage[normalem]{ulem}

\title{\huge{\vspace{-.4in}Data-Driven Aerospace Engineering: \\ Reframing the Industry with Machine Learning}\vspace{-.1in}}
 \author{
   Steven L. Brunton$^1$\footnote{Corresponding author: sbrunton@uw.edu}\,,  J. Nathan Kutz$^2$, Krithika Manohar$^1$,\\
    Aleksandr Y. Aravkin$^2$, Kristi Morgansen$^3$\\ 
   Jennifer Klemisch$^4$, Nicholas Goebel$^4$, James Buttrick$^5$\\
   Jeffrey Poskin$^6$, Agnes Blom-Schieber$^5$, Thomas Hogan$^6$, Darren McDonald$^4$\vspace{.05in}\\
\vspace{-.05in}
\small  $^1$ Mechanical Engineering, University of Washington, Seattle, WA 98195\\\vspace{-.05in}
\small  $^2$ Applied Mathematics, University of Washington, Seattle, WA 98195\\\vspace{-.05in}
\small  $^3$ Aeronautics and Astronautics, University of Washington, Seattle, WA 98195  \\\vspace{-.05in}     
\small  $^4$ Boeing Test \& Evaluation, The Boeing Company, Seattle, WA 98108\\\vspace{-.05in}
\small  $^5$ BCA Engineering, The Boeing Company, Seattle, WA 98108\\\vspace{-.05in}
\small  $^6$ Boeing Research \& Technology, The Boeing Company, Seattle, WA 98108\vspace{-.2in}}
 \date{}

\begin{document}

\maketitle

\begin{abstract}
Data science, and machine learning in particular, is rapidly transforming the scientific and industrial landscapes.  
The aerospace industry is poised to capitalize on big data and machine learning, which excels at solving the types of multi-objective, constrained optimization problems that arise in aircraft design and manufacturing.  
Indeed, emerging methods in machine learning may be thought of as \emph{data-driven} optimization techniques that are ideal for high-dimensional, non-convex, and constrained, multi-objective optimization problems, and that improve with increasing volumes of data.  
In this review, we will explore the opportunities and challenges of integrating data-driven science and engineering into the aerospace industry.  
Importantly, we will focus on the critical need for interpretable, generalizeable, explainable, and certifiable machine learning techniques for safety-critical applications.  
This review will include a retrospective, an assessment of the current state-of-the-art, and a roadmap looking forward. 
Recent algorithmic and technological trends will be explored in the context of critical challenges in aerospace design, manufacturing, verification, validation, and services.  
In addition, we will explore this landscape through several case studies in the aerospace industry.  
This document is the result of close collaboration between UW and Boeing to summarize past efforts and outline future opportunities.  
\end{abstract}
\vspace{0. in}

\section{Introduction}\label{Sec:Intro}
\begin{figure}[t!]
    \centering
    \vspace{-.05in}
    \includegraphics[width=\textwidth]{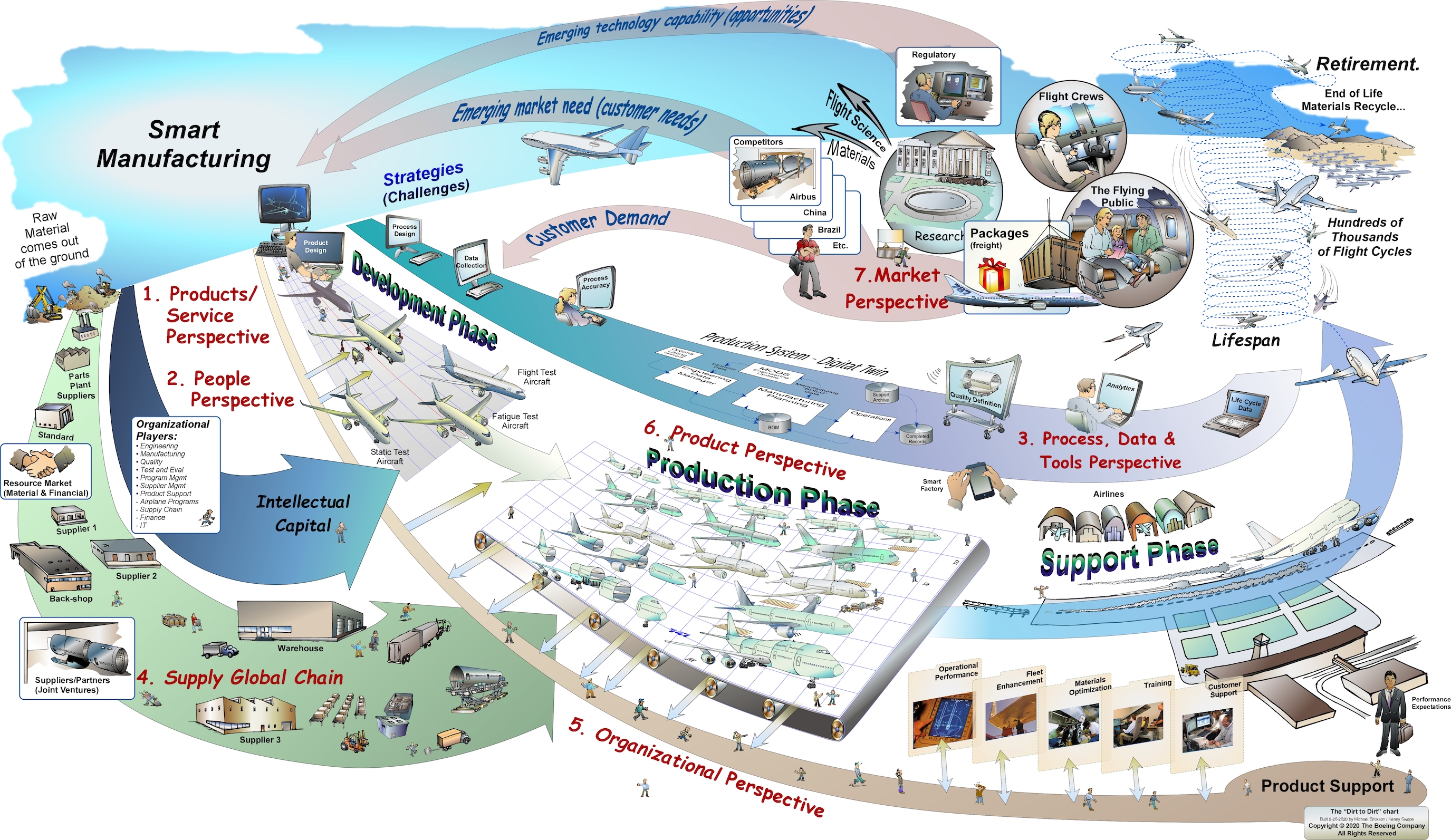}
    \vspace{-.275in}
    \caption{Data science and machine learning have the potential to revolutionize the aerospace industry.}    \label{Fig:Dirt2Dirt}     
    \vspace{-.05in}
\end{figure}

Data science is broadly redefining the state-of-the-art in engineering practice and what is possible across the scientific, technological, and industrial landscapes. 
The big data era mirrors the scientific computing revolution of the 1960s, which gave rise to transformative engineering paradigms and allowed for the accurate simulation of complex, engineered systems.
Indeed, scientific computing enabled the prototyping of aircraft design through physics-based emulators that resulted in substantial cost savings to aerospace manufacturers. 
The Boeing 777 was the first aircraft to have been designed completely from simulation without a mock-up.
In a similar fashion, machine learning (ML) and artificial intelligence (AI) algorithms are ushering in one of the great technological developments of our generation~\cite{harding2006data,Lynch2008nature,Wu2008kis,Marx2013nature,Khoury2014science,Einav2014science,Jordan2015science}.   
The success of ML/AI has been undeniable in traditionally challenging fields, such as machine vision and natural language processing, fraud detection, and online recommender systems.  
Increasingly, new ML/AI opportunities are emerging in engineering disciplines where processes are governed by \emph{physics}~\cite{Kutz2013book,Brunton2019book}, such as materials science, robotics, and the discovery of governing equations from data.  
An overview of opportunities in data-intensive aerospace engineering is shown in Fig.~\ref{Fig:Dirt2Dirt}. 

Advances in data-driven science and engineering have been driven by the unprecedented confluence of (i) vast and increasing data, (ii) advances in high-performance computation, (iii) improvements to sensing technologies, data storage, and transfer, (iv) scalable algorithms from statistics and applied mathematics, and (v) considerable investment by industry, leading to an abundance of open-source software and benchmark problems. 
Nowhere is the opportunity for data-driven advancement more exemplified than in the field of aerospace engineering, which is data rich and is already built on a constrained multi-objective optimization framework that is ideally suited for modern techniques in ML/AI.  
Each stage of modern aerospace manufacturing is data-intensive, including manufacturing, testing, and  service.  
A Boeing 787 comprises 2.3 million parts that are sourced from around the globe and assembled in an extremely complex and intricate manufacturing process, resulting in vast multimodal data from supply chain logs, video feeds in the factory, inspection data, and hand-written engineering notes.  
After assembly, a single flight test will collect data from 200,000 multimodal sensors, including asynchronous signals from digital and analogue sensors, including strain, pressure, temperature, acceleration, and video.  
In service, the aircraft generates a wealth of real-time data, which is collected, transferred, and processed with 70 miles of wire and 18 million lines of code for the avionics and flight control systems alone.  
Thus, big data is presently a reality in modern aerospace engineering, and the field is ripe for advanced data analytics with machine learning.  

The use of data for science and engineering is not new~\cite{Donoho2015data}, and most key breakthroughs in the past decade have been fundamentally catalyzed by advances in data quality and quantity.  
However, with an unprecedented ability to collect and store data~\cite{Hey2003data}, we have entered a new era of \emph{data-intensive} analysis, where hypotheses are now driven by data. 
This mode of data-driven discovery is often referred to as the \emph{fourth paradigm}~\cite{Hey2009msr}, which does not supplant, but instead complements the established modes of theoretical, experimental, and numerical inquiry. 
In fact, we again emphasize the strong parallel with the rise of computational science in the past half century, which did not replace, but instead augmented theory and experiments. 
Just as today computational proficiency is expected in the workforce, so will data science proficiency be expected in the future.  
Thus, the data-intensive transformation in the aerospace industry can learn from the digital transformation over the past decades.  
Perhaps the largest change will be in how teams of researchers and engineers are formed with domain expertise and essential data science proficiency, along with changes in research and development cycles for industry.  

\begin{figure}[t!]
    \centering
    \includegraphics[width=.475\textwidth]{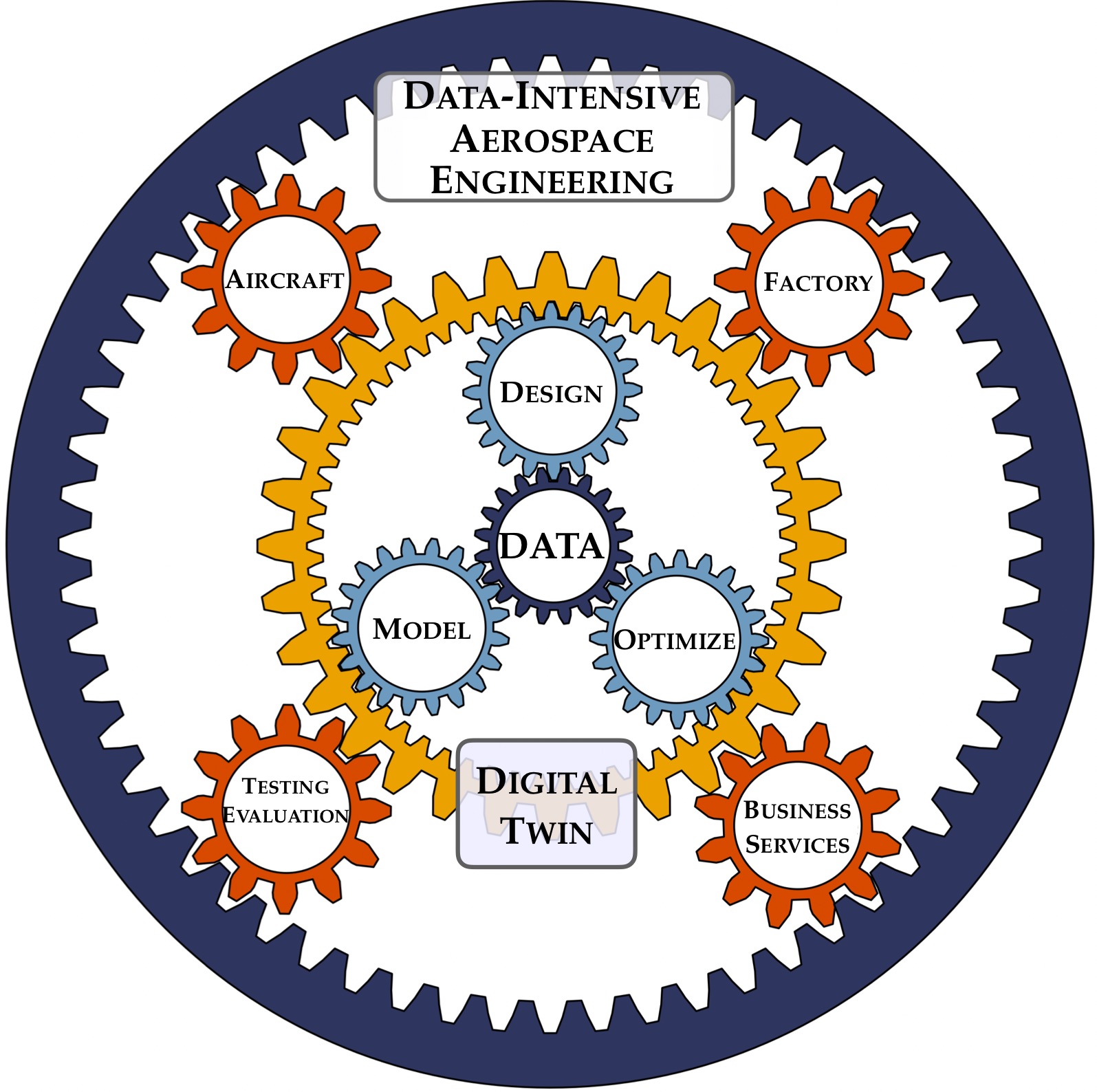}
    \vspace{-.1in}
    \caption{Schematic overview of data-driven aerospace engineering. }
    \label{Fig:Overview}
\end{figure}

The aerospace industry presents a number of unique opportunities and challenges for the integration of data-intensive analysis techniques and machine learning.  
The transformative impact of data science will be felt across the aerospace industry, including: {\bf (i) in the factory} (design for manufacturability, re-use and standardization, process control, safety, productivity, reproducibility, inspection, automation, drilling, shimming), {\bf (ii) in testing and evaluation} (streamlining testing, certification, anomaly detection, data-driven modeling), {\bf (iii) in the aircraft} (inspection, design and performance, materials and composites, maintenance, future product development), {\bf (iv) in human-machine interactions} (advanced design interfaces, interactive visualizations, natural language interactions) and {\bf (v) in the business} (supply chain, sales, human resources, and marketing). 
Because of the exacting tolerances required in aerospace manufacturing, many of these high-level objectives are tightly coupled in a constrained multi-objective optimization. Traditionally, this optimization has been too large for any one group to oversee, and instead, individual components are optimized locally within acceptable ranges. However, unforeseen interactions often cause considerable redesign and program delays.  In the worst-case scenario, accidents may occur.  With improvements in end-to-end database management and interaction (data standardization, data governance, a growing data-aware culture,  and system integration methods), it is becoming possible to create a \emph{digital thread} of the entire design, manufacturing, and testing process, potentially delivering dramatic improvements to this design optimization process.  Further, improvements in data-enabled models of the factory and the aircraft, the so-called \emph{digital twin}, will allow for the accurate and efficient simulation of various scenarios.  In addition to these operational improvements, advances in data-intensive analysis are also driving fundamental advances in aerospace critical fields such as fluid mechanics~\cite{Duraisamy2019arfm,Brunton2020arfm} and material science~\cite{brunton2019methods}.    Importantly, data science works in concert with existing methods and workflows, allowing for transformative gains in predictive analytics and design insights gained directly from data.  Figure~\ref{Fig:Overview} provides a schematic of this process. 

Despite these tremendous potential gains, there are several challenges facing the integration of data-science in the aerospace industry.  
Foremost, due to the safety-critical aspect of aerospace engineering, data-driven models must be certifiable and verifiable, must generalize beyond the training data, and must be both interpretable and explainable by humans.  
Further, collecting vast amounts of data may lead to a \emph{data mortgage}, where simply collecting and maintaining this data comes at a prohibitively high cost, and the data is cumbersome for downstream analysis.  
In the aerospace industry, customers are extremely diverse, with training and certification often done by a range of companies and regulatory authorities, respectively, motivating a level of robustness that is typically not required in other industries. 
There are also fundamental differences between commercial and defense sectors, limiting the sharing of designs, transfer of technology, and joint testing; in fact, at Boeing, testing and evaluation of commercial and defense vehicles have only recently been grouped under one organization, and these efforts still remain siloed.  

The aerospace industry has always been a leader in optimization, and the earliest advances of the Wright brothers may be viewed as an optimization of the flight control system. 
In the century that followed, much of the aerospace industry has been centered around a constrained, multi-objective optimization with an exceedingly large number of degrees of freedom and nonlinear interactions.  
ML algorithms are a growing set of data-intensive optimization and regression techniques that are ideal for these types of high-dimensional, nonlinear, non-convex, and constrained optimizations.  
Aided by advances in hardware and algorithms, modern ML is poised to enable this optimization, allowing a much broader and integrated perspective.  
It is important to note that not all ML is \emph{deep learning} or \emph{artificial intelligence}.  
ML is simply optimization that is directed on data rather than first principles models, so that it fits naturally into existing efforts, but leverages a growing and diverse set of data.

In this paper, we will explore the evolving landscape of data-driven aerospace engineering.  
We will discuss emerging technology and how these are changing what is possible in aerospace design, manufacturing, testing, and services.  
These advances will be viewed through case studies, providing a review of past work and a roadmap for the future. 
Further, we wish to temper the excitement about the immense opportunities of data science with a realistic view of what is easy and what is hard.  
Finally, because the field is evolving rapidly, we want to establish a common terminology, taxonomy, and hierarchy for technology surrounding data science and its ML algorithms.  

\section{Machine Learning and Optimization}\label{Sec:MLOptimization}
In this section, we will discuss a number of mathematical architectures that are central to the data-intensive modeling of aerospace systems.   
First, we will focus on understanding modern machine learning algorithms that will be critical to process data from the aerospace industry.  
We then provide a brief overview of applied optimization, which is the mathematical underpinning of machine learning.  
Next, we discuss specific machine learning extensions and considerations for systems governed by physics, and also discuss the importance of scalable and robust numerical algorithms.  
The treatment of these mathematical topics is far from comprehensive~\cite{Brunton2019book}, and is instead intended to provide a brief, high-level overview.  
Potential uses for these algorithms will be explored in later sections.  

\subsection{Machine learning}
\begin{figure}[t!]
    \centering
    \includegraphics[width=\textwidth]{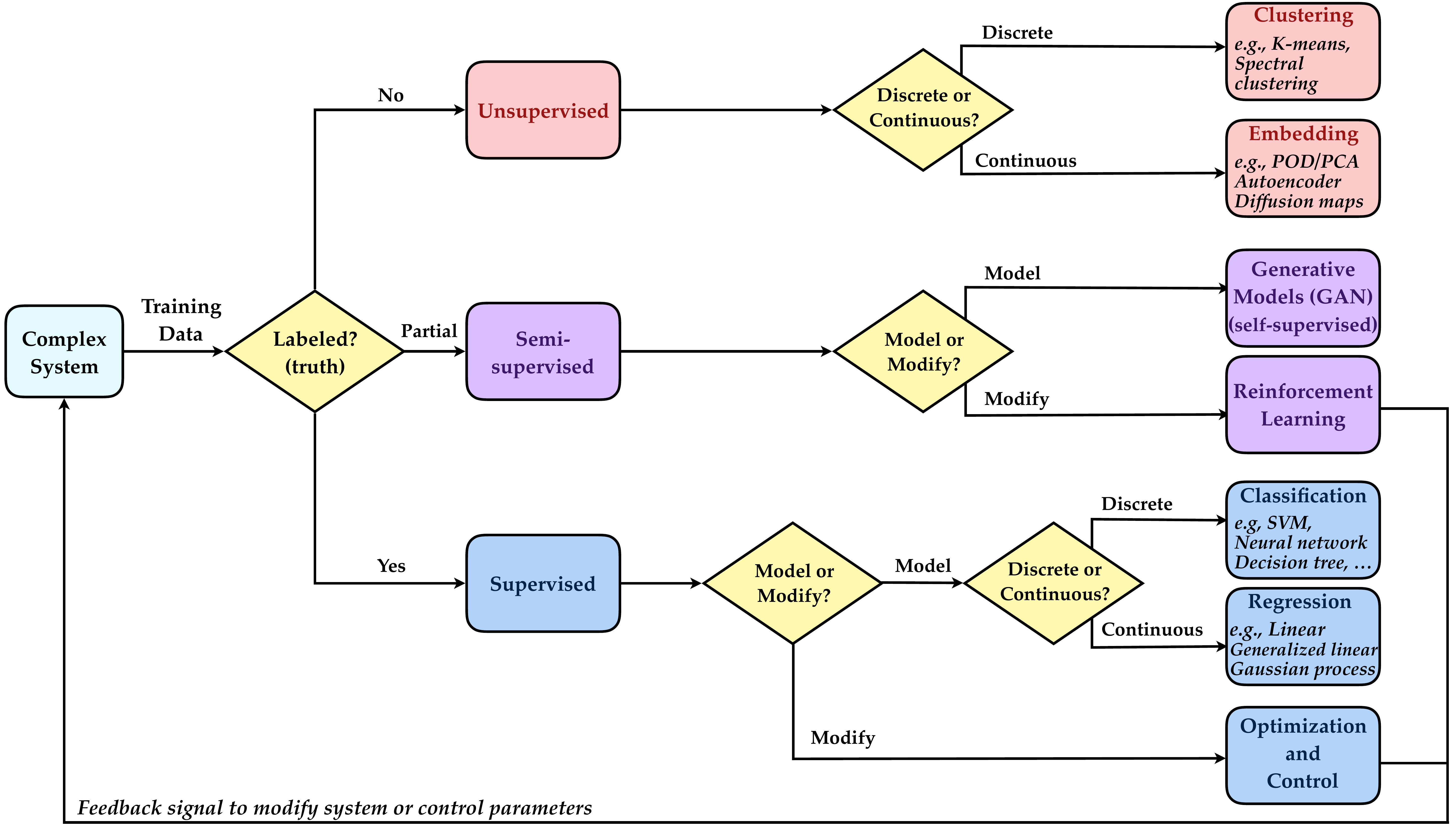}
    \caption{Schematic overview of various machine learning techniques. }
    \label{Fig:MLSchematic}
\end{figure}

Machine learning is a growing set of optimization and regression techniques to build models from data.  
There are a number of important dichotomies with which we may organize the variety of machine learning algorithms. 
Here, we will group these into \emph{supervised} and \emph{unsupervised} learning methods, based on the extent to which the training data is labeled. 
An approximate organization of these learning techniques, organized by task, is shown in Fig.~\ref{Fig:MLSchematic}.  
For example, in supervised approaches, a function $\boldsymbol{\phi}$ mapping input data $\mathbf{x}$ to outputs $\mathbf{y}$, also known as labels or targets, must be learned.  
Typically, the mapping may either predict the label 
\begin{align}
\hat{\mathbf{y}} = \boldsymbol{\phi}(\mathbf{x}; \boldsymbol{\theta})
\end{align}
or model the joint probability distribution of the inputs and outputs
\begin{align}
\boldsymbol{\phi}(\mathbf{x}, \mathbf{y}; \boldsymbol{\theta}).
\end{align}
In both cases, the models are parameterized by the parameters $\boldsymbol{\theta}$, which must be learned.  
There are numerous techniques to learn the structure and parameters of these mappings, including linear and nonlinear regression, genetic programming (i.e., symbolic regression), and neural networks.  
In addition to these dichotomies of ML, \emph{reinforcement} learning, or \emph{semi-supervised} learning, provides yet another paradigm for learning where delayed reward structures can be achieved for tasks such as autonomy (self-driving cars, UAVs, etc).

There are four major stages in machine learning: 1) determining a high-level task or objective, 2) collecting and curating the training data, 3) identifying the model architecture and parameterization, and 4) choosing an optimization strategy to determine the parameters of the model from the data.  Human intelligence is critical in each of these stages.  Although considerable attention is typically given to the learning architecture, it is often the data collection and optimization stages that require the most time and resources.  Indeed, exploring a broad and diverse set of ML architectures is a hallmark feature of today's ML-centric companies. 
 It is also important to note that known physics (e.g., invariances, symmetries, conservation laws, constraints, etc.) may be incorporated in each of these stages.  For example, rotational invariance is often incorporated by augmenting the training data with rotated copies, and translational invariance is often captured using convolutional neural network architectures. { In kernel-based techniques, such as Gaussian process regression and support vector machines, symmetries can be imposed by means of rotation-invariant, translation-invariant, and symmetric covariance kernels.} Additional physics and prior knowledge may be incorporated as additional loss functions or constraints in the optimization problem.  

\subsubsection{Supervised learning}
Supervised learning assumes that the training data $\mathbf{x}$ has labels $\mathbf{y}$.  If the labels are discrete, such as a categorical description of an image (e.g., dog vs. cat), then the supervised learning task is a \emph{classification}.  If the labels are continuous, such as the lift profile for a particular airfoil shape, then the task is a \emph{regression}.  
In the simplest form, the goal of supervised learning is to train a model to minimize the loss function 
\begin{align}
L = \| \mathbf{y}-\boldsymbol{\phi}(\mathbf{x};\boldsymbol{\theta}) \|
\end{align}
where $\|\cdot\|$ is the root mean-squared error (RMSE).  

Labeling the data with expert knowledge often makes it possible for supervised learning algorithms to automate complex tasks.  
Many of the dominant, industrially used algorithms are supervised in nature, including the ubiquitous methods of classification trees and support vector machines~\cite{Brunton2019book,Wu2008kis}.  
More recently, neural networks have surpassed the performance of these methods, provided a sufficiently large data set is available to train the network~\cite{Goodfellow2016book}. 
In supervised learning, the availability of an extensive, high-quality labeled data set is crucial, as in the 2009 ImageNet data set~\cite{deng2009imagenet}, which enabled the training of a deep convolutional neural network that outperformed all previous architectures~\cite{Krizhevsky2012nips}. 

\subsubsection{Unsupervised learning}
Unsupervised learning, also known as data mining or pattern extraction, determines the underlying structure of a data set without labels.  Again, if the data is to be grouped into distinct categories, then the task is clustering, while if the data has a continuous distribution, the task is an embedding.  Unsupervised learning is an extremely challenging task as the algorithm is unguided by expert labels.   Such algorithms are commonly used in an exploratory fashion to learn about data and the kinds of correlations that exist between features.  The three most commonly used methods for unsupervised clustering include $k$-means, mixture models, and hierarchical clustering~\cite{Brunton2019book,Wu2008kis}.  In each case, the number of distinct patterns in the data is usually specified by the user and refined as a tuned hyper-parameter.  
Learning features in an unsupervised manner can often lead to future developments that are supervised. 

One of the standard approaches in embedding is to find a low-dimensional subspace or submanifold, parameterized by a latent variable $\mathbf{z}$, that describes a high-dimensional state $\mathbf{x}$.  In this case, the goal is to find two functions, an encoder $\mathbf{z} =\boldsymbol{\varphi}(\mathbf{x})$ and a decoder $\hat{\mathbf{x}}=\boldsymbol{\psi}(\mathbf{z})$, so that $\hat{\mathbf{x}} = \boldsymbol{\psi}(\boldsymbol{\phi}(\mathbf{x}))\approx \mathbf{x}$.  The functions $\boldsymbol{\varphi}$ and $\boldsymbol{\psi}$ are implicitly parameterized by weights $\boldsymbol{\theta}$ that must be tuned to minimize the following loss function:
\begin{align}
L = \|\mathbf{x} - \boldsymbol{\psi}(\boldsymbol{\phi}(\mathbf{x}))\|.
\end{align}
When the encoder and decoder are linear functions, given by matrices, then the optimal embedding recovers the classical singular value decomposition (SVD) or principal component decomposition (PCA)~\cite{Brunton2019book,Brunton2020arfm}.  
However, these functions may be nonlinear, defined by neural networks, resulting in powerful autoencoders.

\subsubsection{Reinforcement learning}
The power of reinforcement learning (RL)~\cite{sutton2018reinforcement} lies in its ability to learn from interactions with the environment with goal-oriented objectives.  Thus its application domain includes autonomy and control.  This is unlike the two other dominant ML paradigms of supervised and unsupervised learning~\cite{Goodfellow2016book,Brunton2019book}.  With a trial-and-error search, an RL agent senses the state of its environment and learns take appropriate actions to achieve optimal immediate or delayed rewards. Specifically, the RL agent arrives at different states $\mathbf{s}$ by performing actions $\mathbf{a}$, with the selected actions leading to positive or negative rewards $\mathbf{r}$ for learning.  Importantly, the RL agent is capable of learning delayed rewards, which is critical for many systems where a circuitous path to the optimal solution must be learned.  
Rewards may be thought of as sporadic and delayed labels, leading to RL often being classified as \emph{semi-supervised} learning.  
One canonical example is learning a set of moves, or a long term strategy, to win a game of chess.  

Reinforcement learning is often formulated as an optimization to determine the policy $\boldsymbol{\pi}(\mathbf{s},\mathbf{a})$, which is a probability of taking action $\mathbf{a}$ given state $\mathbf{s}$, to maximize the total reward across an episode.   Given a policy $\boldsymbol{\pi}$, it is possible to define a value function that quantifies the desirability of being in a given state:
\begin{align}
V_{\boldsymbol{\pi}}(\mathbf{s}) = \mathbb{E}\left(\sum_t \gamma^t \mathbf{r}_t | \mathbf{s}_0=\mathbf{s}\right),
\end{align}
where $\mathbb{E}$ { is the expected reward over the time steps $t$, subject to a {\em discount rate} $\gamma$.} 
Typically, it is assumed that the state evolves according to a Markov decision process, so that the probability of the system occurring in the current state is determined only by the previous state.   Thus a large number of trials must be evaluated in order to determine an optimal policy.  This is accomplished in chess and Go by self-play~\cite{silver2018general}, which is exactly what many engineered systems are allowed to do to learn an optimal policy.  
Often, in modern deep reinforcement learning, a deep neural network is used to learn a \emph{quality} function $Q(\mathbf{s},\mathbf{a})$ that jointly describes the desirability of a given state/action pair.

\subsubsection{Deep learning}
Deep learning, or learning based on neural networks (NNs) with a deep multi-layer structure, is often synonymous with ML and is the core architecture for many modern supervised and reinforcement learning paradigms.  
Neural networks are particularly powerful due to their expressive representations of data and their diverse architectures~\cite{Goodfellow2016book}. 
The unparalleled success of these algorithms in ML is due to the availability of sufficiently vast and rich training data and modern computational hardware, which have enabled the training of exceedingly large neural networks with millions or billions of free parameters.  

\begin{figure}[t]
\vspace*{-1.2in}
\hspace*{-1.5in}
    \begin{overpic}[width=1.3\textwidth]{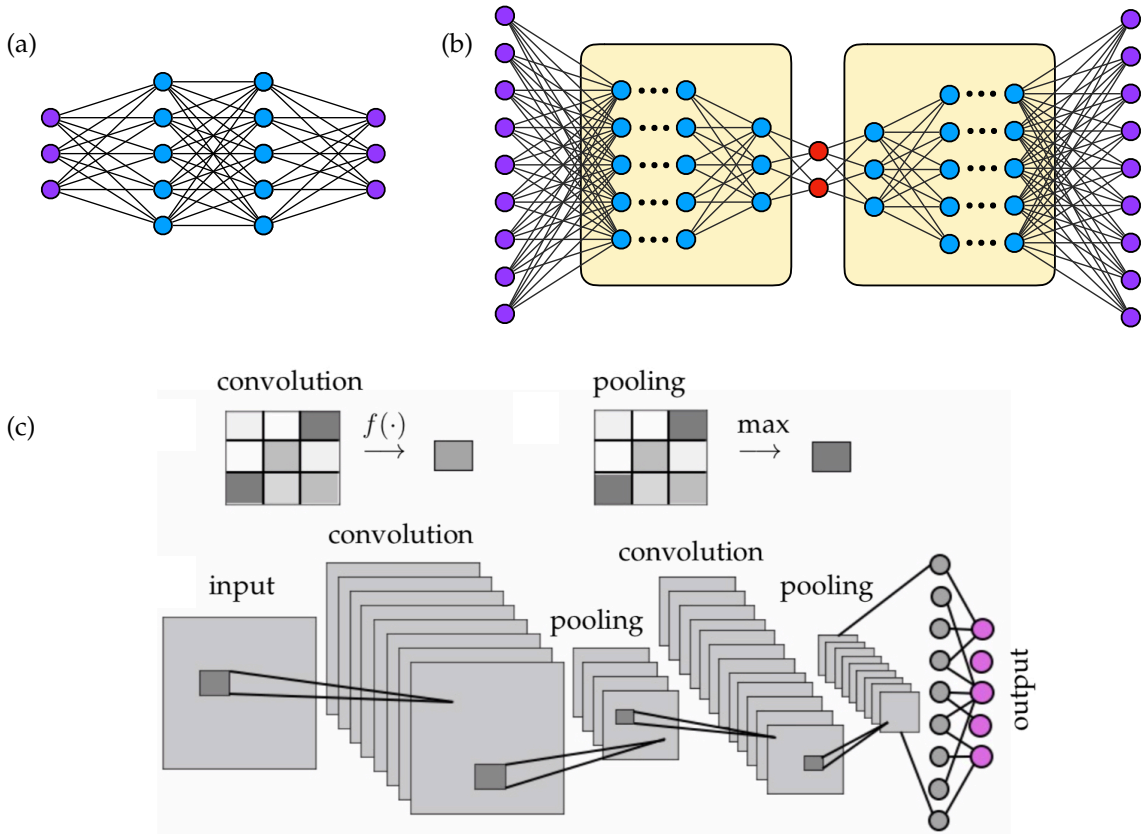}
    \put(20,60){(a)}
    \put(45,60){(b)}
    \put(20,38){(c)}
    \end{overpic}
\vspace*{-1.45in} 
    \caption{Mathematical architectures of commonly used neural networks, including (a) a feed-forward neural network, (b) a deep autoencoder network, and (c) a deep convolutional neural network.  }
    \label{Fig:NN1}
\end{figure}

Figure~\ref{Fig:NN1} shows a number of the leading architectures used in practice today in a variety of applications, including a simple feedforward architecture, a deep autoencoder that capitalizes on low-dimensional structure in data, and a deep convolution NN (DCNN) that is widely used to extract features for machine vision.  
The term deep refers to the number of neural network layers (typically 7-10) used to map from inputs to outputs.  Neural networks are universal function approximators~\cite{hornik1989multilayer}, and they assume a compositional structure
\begin{align}
\mathbf{y} = {\boldsymbol{\phi}}_1 ({\boldsymbol{\phi}}_2(  \cdots ({\boldsymbol{\phi}}_n(\mathbf{x};{\boldsymbol{\theta}}_n);\cdots );\boldsymbol{\theta}_2);\boldsymbol{\theta}_1).
\end{align}
The flexibility of this compositional structure enables the construction of classification or regression maps between input and output data.  With sufficient data, one can optimize for the NN weight parameters $\boldsymbol{\theta}_n$, usually via some form of stochastic gradient descent.    
Deep learning is commonly used in commercial settings, with DCNNs being the state-of-the-art for characterizing images and spatial correlations, and recurrent NNs (RNNs) enabling powerful speech and text recognization software.  NNs are typically supervised learners and require a significant amount of data.  They are also known to overfit to data and fail to generalize to new parameter regimes.  Regardless, they are a powerful technology that can power many of the supervised learning tasks required in modern data science applications.

\subsubsection{Physics informed learning}
Physics informed learning~\cite{Loiseau2017jfm,Raissi2017arxiv,Loiseau2018jfm,battaglia2018relational,raissi2019physics,noe2019boltzmann,kohler2019equivariant,cranmer2019learning,raissi2020hidden,cranmer2020discovering,cranmer2020lagrangian,Brunton2020arfm} is of growing importance for scientific and engineering problems.  
\textit{Physics informed} simply refers to our ability to constrain the learning process by physical and/or engineering principles.  For instance, conservation of mass, momentum, or energy can be imposed in the learning process~\cite{Loiseau2017jfm}.  In the parlance of ML, the imposed constraints are referred to as regularizers.  Thus, physics informed learning focuses on adding regularization to the learning process to impose or enforce physical priors.  For the example of a NN model, this becomes
\begin{align}
 \argmin_{\boldsymbol{\theta}_1, \boldsymbol{\theta}_2, \cdots, \boldsymbol{\theta}_n} 
\mathbf{y} = \boldsymbol{\phi}_1 (\boldsymbol{\phi}_2(  \cdots (\boldsymbol{\phi}_n(\mathbf{x};\boldsymbol{\theta}_n);\cdots );\boldsymbol{\theta}_2) \boldsymbol{\theta}_1) + \lambda g(\boldsymbol{\theta}_1, \boldsymbol{\theta}_2, \cdots, \boldsymbol{\theta}_n) ,  
\end{align}
where the regularization $g(\cdot)$ imposes the desired physical constraint.  The parameter $\lambda$ is a hyper-parameter allowing the user to impose an increasingly strong regularization to enforce this constraint.  The importance of this constraint, or potentially multiple constraints, cannot be overstated in engineering and physics systems.  
Specifically, this is where known physics or physical constraints can be explicitly incorporated into the data-driven modeling process.   Physics informed learning, often enacted with deep learning architectures, represents the state-of-the-art in ML methods for the engineering and physical sciences.

Rather than imposing physical constraints explicitly, an alternative is to learn embeddings based on physical models. 
This physics-guided paradigm involves learning embeddings from data produced by known first principles models of physics.
The computation of these embeddings, in the context of aerospace and fluid dynamics, is often known as {\em modal analysis}, and has become increasingly data-driven (either from simulation or observation) in recent years~\cite{Taira2017aiaa,taira2020modal,Brunton2020arfm}.
The physical coupling between fluids and aerospace structures are particularly important and the modes of these coupled interactions are impossible to discern by analyzing the Navier-Stokes equations and structural models alone -- instead, they are determined by the boundary interactions between the coupled models.
Thus, this physics guided architecture greatly enhances the understanding and design of robust engineering systems that can withstand complex interactions, turbulence, and instabilities.  Improved reduced-order models of fluid dynamics may further aid efforts in flow control~\cite{colonius2001overview,Brunton2015amr} and to reduce jet noise~\cite{Jordan2013arfm}.   
Modern modal analysis techniques, such as POD/PCA~\cite{Brunton2019book}, dynamic mode decomposition~\cite{Schmid2010jfm,Tu2014jcd,Kutz2016book}, Koopman mode decomposition~\cite{Mezic2013arfm,Brunton2016plosone}, and resolvent mode analysis~\cite{mckeon2010critical}, naturally fall under the umbrella of unsupervised learning. 
Furthermore, the dimensionality reduction achieved by these methods enable low-latency, efficient downstream tasks such as system identification~\cite{Brunton2016pnas}, airfoil shape optimization and uncertainty quantification~\cite{renganathan2020koopman}, and reduced-order modeling~\cite{Noack2003jfm,Carlberg2013jcp,bui2007goal,bui2008model,amsallem2015design,Benner2015siamreview,Carlberg2017jcp,carlberg2018recovering,singh2017machine,taira2020modal}.

\subsection{Optimization}
All of machine learning relies on optimization~\cite{Brunton2019book,Brunton2020arfm}.  
In fact, ML may be viewed as a growing set of applied optimization algorithms to build models from data.  
Mathematical optimization~\cite{Boyd2004convexbook} comprises three interconnected areas: theoretical underpinnings, algorithm design and implementation, and modeling. 
Modeling allows us to bring in deterministic or physical descriptions of the real-world, along with information about structure of unknowns and statistical measures that describe beliefs about error and uncertainty. 
The scope of problems that can be formulated in this way is broad, ranging from 
sensing, estimation, and control~\cite{Kalman1960jfe,bertsekas1995dynamic,rao1998application,dp:book,sp:book,Joshi2009ieee,aravkin2017generalized,Manohar2017csm} to  
machine learning~\cite{Goodfellow2016book,bottou2018optimization,Brunton2019book} and decisions under uncertainty~\cite{shapiro2009lectures}. 
In all cases a modeling process gives rise to an optimization problem, where minimizing and maximizing over parameters leads to the desired inference or learning machine.  Theoretical developments, such as convex and variational analysis, capture properties of problem formulations such as smoothness, convexity, and well-posedness of the problems themselves.  In addition, deriving provable behavior of such algorithms is often important to guarantee performance.  

A major distinction in optimization is between {\it convex} and {\it nonconvex} problems.  
Here, convexity refers to the property of the objective function to be minimized or maximized and the set of values being optimized over.  
Convex optimization problems~\cite{Boyd2004convexbook} are extremely well studied, as there are fast and scalable generic solution techniques with performance guarantees.  
Convex objective functions will have a single global minima or maxima (i.e., the function has a single \emph{hill} or \emph{valley}), and gradient-based methods may be used to converge to this extremum.  
In contrast, nonconvex models, which have many local minima and maxima (i.e., the objective function has many local hills and valleys), form a much wider, and hence richer, class of problems. 
But, as such, there are no general scalable solution techniques or strong convergence guarantees.  

\subsubsection{Stochastic algorithms} Many problems in machine learning involve training sets with millions of datapoints, making standard gradient computations prohibitively expensive. 
Stochastic optimization algorithms use random sampling to scale gradient descent algorithms to such data sets by using small subsets of the data at any given time. 
While these methods have a long history~\cite{tsitsiklis1986distributed}, recent developments have focused on extending these ideas to machine learning~\cite{bottou2010large,kingma2014adam}. 
Recent theoretical breakthroughs have shown that the simplest algorithms that directly use sampled 
gradients, and even approximate gradients (subgradients) in the nonsmooth case, are provably convergent for a far wider problem class than was previously known~\cite{davis2018stochastic}.  
These results justify the prolific use of these methods for general large-scale problems, such as training neural networks.
In addition, stochastic algorithms for structured nonconvex and nonsmooth problems make heavy use of the proximity operator~\cite{reddi2016proximal,aravkin2019trimmed}, 
and can converge more rapidly than classic stochastic methods by exploiting variance reduction techniques to obtain improved search directions. 

\subsubsection{The role of structure in algorithm design} A key theme in the field of optimization is to find the right problem classes that balance general applicability with specificity for the design of efficient algorithms. 
Rather than thinking of problems in generality such as convex versus nonconvex, more specific assumptions, reminiscent of physics-informed learning constraints, allow faster algorithms and better guarantees. 
Within convex optimization, the piecewise linear-quadratic class~\cite{rockafellar2009variational} captures a wide range of models and admits specialized solution techniques. 
Outside of convex optimization, the class of convex-composite~\cite{burke1995gauss} problems is a 
key generalization that has seen recent algorithmic development and analysis~\cite{davis2019stochastic}. 
Coupled problems with multiple parameters have been solved for numerous applications with variable projection techniques~\cite{golub2003separable}.  
Nonconvex composite models and algorithms have also been developed, with many applications involving challenging data-generating mechanisms and sparse regularization~\cite{Zheng2019ieeeacess}.

\subsubsection{Atomic operations for nonconvex nonsmooth functions}

There have been significant recent advances in nonconvex, nonsmooth optimization where the objective functions are not differentiable, and where parameters are constrained, for example by physical bounds. Neural networks and deep learning models have been a particularly influential driver of methods for large-scale nonconvex models. 
Complementing this direction, nonsmooth models frequently arise as a means of imposing structure on the solution, for example sparsity.  
In the remainder of this section, we discuss some of these advances in more detail with references to survey literature in optimization.

When optimizing a smooth function, algorithms fundamentally rely on gradients to implement second order methods, such as Newton's method or the Gauss-Newton and quasi-Newton variants (see~\cite{nocedal2006numerical} for an overview). 
These gradient computations involve matrix-vector products and equation solves, which we will call {\it atomic operations}. 
Most common algorithms can be decomposed into such operations.  

Another key operation is that of the {\it proximity operator}, which has a long history and a tremendous range of recent applications~\cite{combettes2011proximal,parikh2014proximal}. 
Given a function $f$, we define 
\begin{equation}
\label{eq:prox}
\text{prox}_{\alpha f}(z) = \argmin_{x} \frac{1}{2\alpha} \|x-z\|^2 + f(x).
\end{equation}
In words, we minimize the sum of the function and a scaled quadratic around a base point $z$ and return the minimizing value. 
When the function $f$ is arbitrary, evaluating the operator may be difficult or impossible. However, under moderate assumptions that are satisfied for a wide range of applications, the proximity operator has a closed form solution or a provably fast computational routine. 
Many algorithms for solving nonsmooth nonconvex problems use the proximity operator as a subroutine, and are provably convergent~\cite{attouch2013convergence}.

\subsection{Scalable and robust algorithms}
Despite the growing abundance of measurement data across the engineering sciences, systems are often fundamentally low-dimensional, exhibiting a few dominant features that may be extracted using dimensionality reduction~\cite{Taira2017aiaa,Brunton2019book}.  
The existence of low-rank patterns facilitates efficient models and sparse sampling, as there are only a few important degrees of freedom that must be characterized, regardless of the ambient measurement dimension.  
Here we discuss some of the tremendous advances in the past decades developing robust and scalable numerical algorithms for big data applications.  

\subsubsection{Randomized linear algebra}\label{Sec:Methods:Randomized}

Massive datasets pose a computational challenge for traditional algorithms, placing significant constraints on memory, processing power, and computational time. 
Recently, the powerful concept of {randomization} has been introduced as a strategy to ease the computational load while still achieving performance that is comparable with traditional matrix factorization techniques.  
The critical idea of probabilistic algorithms is to employ some degree of randomness in order to derive a smaller matrix from a high-dimensional data matrix. The smaller matrix is then used to compute the desired low-rank approximation. Such algorithms are shown to be computationally efficient for approximating matrices with low-rank structure.  Of particular interest are randomized routines for the computation of the singular value decomposition (SVD), (robust) principal component analysis (PCA), and CUR decompositions.  

Several probabilistic strategies have been proposed to find a `good' smaller matrix, and we refer the reader to the surveys~\cite{Mahoney2011,halko2011rand,liberty2013simple,erichson2016randomized} for an in-depth discussion, and theoretical results.
In addition to computing the SVD~\cite{sarlos2006improved,Martinsson201147} and PCA~\cite{rokhlin2009randomized,halko2011algorithm}, it has been demonstrated that this probabilistic framework can also be used to compute the pivoted QR decomposition~\cite{doi:10.1137/15M1044680}, the pivoted LU decomposition~\cite{shabat2016randomized}, the CP tensor decomposition~\cite{erichson2020randomized}, and the dynamic mode decomposition~\cite{erichson2019randomized}. It also helps frame computationally tractable reduced order models~\cite{alla2019randomized,bai2020dynamic}.  It should be noted that the tech giants, such as Google and Facebook, routinely use randomized algorithms to analyze their large data sets.

\subsubsection{Robust dimensionality reduction}\label{Sec:Methods:RPCA}
Robust statistical methods are essential for evaluating real-world data, as advocated by John W. Tukey in the early days of data science~\cite{Huber2002as,Donoho2015data}.  
Many techniques in dimensionality reduction are based on least-squares regression, which is susceptible to outliers and corrupted data.  
Principal component analysis suffers from the same weakness, making it \emph{fragile} with respect to outliers.  
To address this sensitivity, Cand\'{e}s et al.~\cite{rpca} introduced a robust PCA (RPCA) that decomposes a data matrix $\bX$ into a  low-rank matrix $\bL$ containing dominant coherent structures, and a sparse matrix $\bS$ containing outliers and corrupt data:
\begin{equation}
\bX = \bL + \bS.
\end{equation}
The principal components of $\bL$ are \emph{robust} to the outliers and corrupt data in $\bS$.  
the low-rank matrix $\bL$ is decomposed via the SVD into $\bL = \bPhi\mathbf{D}\bV^T$, where coherent features are given by the matrix $\bPhi$.  
We generally use the first $r$ dominant columns of $\bPhi$, corresponding to the $r$ features that explain the most variance in the data.  
The SVD provides the best rank-$r$ least squares approximation for a given rank $r$: $\hat\bL = \bPhi_r\mathbf{D}_r\bV_r^T$. The target rank $r$ must be carefully chosen so that the selected features only include meaningful patterns and discard noise corresponding to small singular values. 
The subsequent left singular vectors $\bPhi_r$ are the desired low-rank features that span the columns of $\bL$. 
The truncation parameter $r$ may be determined using the optimal singular value truncation threshold of Gavish and Donoho~\cite{gavish2014optimal}.

RPCA has tremendous applicability for modern problems of interest, including video surveillance~\cite{bouwmans2014robust} (where the background objects appear in $\bL$ and foreground objects appear in $\bS$), natural language processing~\cite{kondor2013using}, matrix completion, and face recognition~\cite{wright2009robust}.  
Matrix completion may be thought of in terms of the Netflix prize, where a matrix of preferences is constructed, with rows corresponding to users and columns corresponding to movies.  This matrix is sparse, as most users only rate a handful of movies.  The goal is to accurately fill in missing matrix entries, revealing likely user ratings for movies they haven't seen.
We will demonstrate the use of RPCA for an aircraft shimming application in the case study in Sec.~\ref{Sec:Shimming}.

\section{Digital Twin and Enabling Technologies}\label{Sec:DigitalTwinn}
Several key technologies are necessary to support the design, manufacturing, testing, and service of tomorrow's aerospace products, which will ultimately be enabled by a robust digital twin.  
These enabling technologies include sensors and the internet of things, a comprehensive digital thread, rapid data access and data storage, virtual reality to perform testing, reduced order models and discrepancy models, uncertainty quantification, autonomy, and control.  
This section will review a number of these emerging technologies. 

\subsection{Digital twins}

Digital twin technology promises to revolution the entire manufacturing and engineering design landscape~\cite{boschert2016digital,grieves2017digital,tao2018digital,rasheed2019digital,chinesta2020virtual}.  The goal of the digital twin is to bridge the physical and virtual worlds, providing a proxy environment to simulate, test, and evaluate model designs at a fraction of the cost of real-world implementation.  
The digital twin relies on an accurate, physics-based emulator that characterizes the statics or dynamics of a given system.  
Typically, this model will integrate a hierarchy of multi-physics and multi-fidelity models, which will be continually updated with data streams from the real world.  
To be more mathematically precise, many physics-based models will consist of a system of nonlinear partial differential equations describing the time-space evolution a system.  Such an evolution equation can be represented as follows
\begin{equation}
  {\bf u}_t  =  {\bf N}\left( {\bf u}, {\bf u}_x, {\bf u}_{xx}, \cdots,   x,t ;\boldsymbol{\beta} \right)
  \label{eq:complex}
\end{equation}
where $  {\bf u}$ is the system state, the subscripts denote partial differentiation, and ${\bf N}(\cdot)$ prescribes the generically nonlinear evolution.  
The vector $\boldsymbol{\beta}$ represents the parameters on which the system depends.  
Eq.~\eqref{eq:complex} also requires a set of initial and boundary conditions on a given domain.  
Typically, high-fidelity simulations of this system of equations is computationally involved, and it may be prohibitively expensive to simulate a truly multiscale system, such as the turbulent fluid flow over a full-scale wing at all scales.  Instead, it is often necessary to leverage reduced-order models that capture dominant physical phenomena at a fraction of the cost.  
Of course, some systems are time-independent and some are spatially independent.  In either case, what is critical is that a proxy, physics-based model exists that is capable of informing how a system behaves.  From robots to manufacturing lines, an accurate virtual representation holds tremendous promise for technological advancement.  

The success of digital twin technology centers on accurate virtual representations of the physical world.  For precision manufacturing, for instance, current digital twin technologies do not provide the necessary level of fidelity for a number of processes. What distinguishes digital twins from traditional modeling efforts is the integration of multi-physics systems and components.  Thus the digital twin often represents an entire engineering process versus individual components in the process, for which we may have good models. 
However, the digital twin must ensure end-to-end performance across the multi-physics system, placing stringent requirements on the fidelity of models and how they communicate.  Discrepancy models (discussed below) provide an adaptive framework, whereby the digital twin can continuously learn updated, high-precision physics models over the course of time from its own sensor network.  The integration of data and multi-physics models across a system is a grand challenge, requiring intelligent, robust, and adaptive architectures for learning and control.  Many of the data-driven strategies discussed here are ideally poised to help in building accurate and viable digital twin models.

\subsection{Sensor technology and the internet-of-things}\label{Sec:Sensors}
The aerospace industry generates tremendous volumes of data from a vast array of distributed sensors.  With emerging \emph{internet-of-things} sensing and communication capabilities, this volume of data will only increase.  
To avoid a data mortgage, where the majority of resources are spent collecting and curating data, it is critical that key features are automatically extracted and analyzed in real time through edge computing.  
Thus, the paradigm of \emph{big data} will shift to one of \emph{smart data}. 
It is also important for algorithms to be robust to outliers.  
Outliers may correspond to sensor failures or saturations, although they may also signal important events that should be analyzed more carefully.  
Identifying where to place new sensors will also play a key role in improving efficiency and process control.  

Many complex systems, such as a turbulent fluid or a large aerospace structure, have many degrees of freedom and are mathematically represented as a high-dimensional vector of data resulting from simulations or physical measurements.  
However, high-dimensional data often exhibit low-dimensional patterns, which is the foundation of dimensionality reduction and machine learning.  
The high-dimensional state $\bx\in\reals^n$ may then be efficiently represented in a low-dimensional basis $\bPhi_r\in\mathbb{R}^{n\times r}$, for example via RPCA above, so that $\bx\approx \bPhi_r\ba$, where $\ba$ is a vector that approximates $\bx$ in terms of the first $r$ principal components $\bPhi_r$.  
Often the structure of $\bPhi_r$ is well-characterized from historical data (e.g., a library of human faces, or a set of point cloud scans of a particular aircraft part across several aircraft).  
In this case, the number of measurements required to estimate the full vector $\bx$ may be dramatically reduced from the ambient dimension $n$. 
These sparse sensors can be selected to best identify the coefficients $\ba$ in the basis $\bPhi_r$, thus enabling robust estimation of the high-dimensional state $\bx$.  
There are a number of sparse sensing paradigms, including {gappy} sampling~\cite{Everson1995gappy,Willcox2006compfl}, empirical interpolation methods (EIM)~\cite{Barrault2004crm,Chaturantabut2010siamjsc,drmac2016siam}, including the discrete empirical interpolation method, or \emph{DEIM}, and compressed sensing~\cite{candes2006compressive,donoho2006compressed,Candes2006cpam,Candes2006ieeetit,Candes2006bieeetit,baraniuk2007compressive}.  

Sparse sampling in a tailored basis~\cite{Dhingra2014cdc,Manohar2017csm,Manohar2016jfs} has been widely applied to problems in the imaging sciences as well as to develop reduced order models of complex physics, such as unsteady fluid dynamics.  
Thus, even if we cannot measure the full state $\bx$, it is often possible to estimate the state from $r\ll n$ point measurements in the {\em observation} vector $\by\in\reals^r$ given by $\by = \bC\bx = \bC\bPhi_r\ba$, where $\bC\in\reals^{r\times n}$ is the point measurement operator. 
Then, the reconstruction of the remaining state reduces to a least squares estimation problem for the coefficients
$\hat\ba = (\bC\bPhi_r)^{\dagger}\by,$
where $\dagger$ is the Moore-Penrose pseudoinverse.  
This procedure was first used to reconstruct facial images from a subsampled pixel mask~\cite{Everson1995gappy}.  Importantly, it permits a drastic reduction in computation by solving for $r\ll n$ unknowns. The full state estimate $\hat\bx$ is subsequently recovered using the feature basis: $\hat\bx = \bPhi_r\hat\ba$. 
The accuracy of reconstruction depends on the structure of the basis $\bPhi_r$ and the choice of observations $\bC$.

{The number of observations can be greatly reduced by optimizing the sensors to maximize the accuracy of reconstruction. However, the combinatorial search over all $n\choose r$ possible sensor locations is  computationally intractable even for moderately large $n$ and $r$.}  However, there are convex relaxations that can be solved using standard optimization techniques and semidefinite programs in $\mathcal{O}(n^3)$ runtime. 
{We advocate a greedy matrix volume maximization scheme using the pivoted matrix QR factorization~\cite{drmac2016siam,Manohar2017csm}. This algorithm treats the selected measurements as optimal rows of the linear operator $\bC\bPhi_r$, which designs sampled features $\bC\bphi_i$ to be as orthogonal to each other as possible. Since the row selected columns $\bphi_i$ are no longer orthonormal, greedy row selection methods attempt to maintain near-orthonormality of the features $\bC\bphi_i$ for a numerically well-conditioned inverse.}

\subsection{Reduced-order modeling}
The engineering sciences increasingly rely on simulations as proxies for modeling expensive experimental systems. 
The complexity and dimension of these numerical simulations are growing rapidly due to increasing computational power and resolution in numerical discretization schemes. 
Many complex PDEs such as Eq.~\eqref{eq:complex}, which are critical for digital twin technologies, yield discretized systems of differential equations with millions or billions of degrees of freedom.  These large systems, such as turbulent fluid flows, are extremely demanding, and may be prohibitively expensive, even for the most advanced supercomputers.  
Yet most dynamics of interest are known to be low-dimensional in nature, in contrast to the high-dimensional nature of scientific computing. Reduced-order models (ROMs) help reduce the computational complexity required to solve large-scale engineering systems by approximating the dynamics with a low-dimensional surrogate~\cite{maute2001coupled,bui2007goal,bui2008model,amsallem2015design,amsallem2015design,Benner2015siamreview}.   

To aid in computation, the selection of a set of optimal basis modes is critical, as it can greatly reduce the number of differential equations generated.  
Many solution techniques involve the solution of a linear system of size $n$, which generically involves $\mathcal{O}(n^3)$ operations. Thus, reducing $n$ is of paramount importance.  It is possible to approximate the state ${\bf u}$ of the PDE using a Galerkin expansion:
 \begin{equation} {\bf u}(t)\approx \bPhi_r {\bf a}(t)
 \label{eq:podG}
 \end{equation} 
where ${\bf a}(t)\in \mathbb{R}^r$ is the time-dependent coefficient vector and $\bPhi_r$ is an optimal basis of orthogonal columns, typically generated by singular value decomposition; in this case $r\ll n$.  
We then substitute this modal expansion into Eq.~\eqref{eq:complex} and leverage the orthogonality of  $\bPhi_r$ to yield the reduced evolution equations
   \begin{equation}
   \label{eq:pod}
     \frac{d{\bf a}(t)}{dt}=\bPhi_r^T \bL \bPhi_r {\bf a}(t)+ \bPhi_r^T \bN(\bPhi_r {\bf a}(t);\boldsymbol{\beta}).
   \end{equation}
   By solving this small system, the solution of the original high-dimensional 
system can be approximated.  
Of critical importance is evaluating the nonlinear terms in an efficient way, for example using gappy POD or other sparse sampling techniques.  Otherwise, evaluating the nonlinear terms still requires calculations of the high-dimensional function with the original dimension $n$.  In certain cases, such as the quadratic nonlinearity of Navier-Stokes, the nonlinear terms can be computed once in an off-line manner.  However, parametrized systems generally require repeated evaluation of the nonlinear terms as the basis may change with $\boldsymbol{\beta}$.
Regardless, ROMs allow one to approximate an $n$-dimensional system with an $r$-dimensional system, where $r\ll n$, making many computations possible that would otherwise be intractable.
As such, ROMs are transforming high-performance computing by allowing for computational studies that have been intractable in the past. 
Recently, model discovery techniques are also providing low-order models of complex systems~\cite{Bongard2007pnas,Schmidt2009science,Brunton2016pnas,Brunton2017natcomm,Loiseau2017jfm,Loiseau2018jfm}.

\subsection{Discrepancy modeling}

First principles modeling of physical systems has led to significant technological advances across all branches of science.   For nonlinear systems, however,  small modeling errors can lead to significant deviations from the true (measured) behavior. Even in mechanical systems, where the equations are assumed to be well-known, there are often model discrepancies corresponding to nonlinear friction, wind resistance, etc. Discovering models for these discrepancies remains an open challenge for many complex systems. 
There are many reasons why model discrepancies occur~\cite{Bayesiancalibrationofcomputermodels,QuanModelUncertainty}. 
First, there may be measurement noise and exogenous disturbances.  
In this case, the Kalman filter may be thought of as a discrepancy model where the mismatch between a simplified model and observations is assumed to be a Gaussian process~\cite{Kalman1960jfe}.  
Second, the parameters of the system may be inaccurately modeled. 
Even worse, the structure of the model may not be correct, either because important terms are missing or erroneous terms are present.  
This is known as \emph{model inadequacy} or model structure mismatch.  
Other challenges include incomplete measurements and latent variables, delays, and sensitive dependence on initial data in chaotic systems. 

Discrepancy modeling centers on parameter and structural uncertainties.  One can consider the following governing equations for a given engineering system given by Eq.~\eqref{eq:complex}.  In general, there is also an output measurement,  
 \begin{align}
     \boldsymbol{y} = h({\bf u};\boldsymbol{\beta})
 \end{align}
from which the state may be estimated.  
The discrepancy modeling problem evaluates the difference between the model output of a quantity of interest $\boldsymbol{\phi}_m(t)$ and the observed value $\boldsymbol{\phi}_o(t)$:
\begin{align}
    \delta\boldsymbol{\phi}(t) = \boldsymbol{\phi}_o(t) - \boldsymbol{\phi}_m(t),
\end{align}
where $\delta\boldsymbol{\phi}$ is the discrepancy. 
The goal of discrepancy modeling is then to characterize the discrepancy $\delta\boldsymbol{\phi}(t)$, for example, using standard {\em Gaussian process regression}~\cite{quinonero2005unifying}, dynamic mode decomposition~\cite{Kutz2016book} for approximating $\delta\boldsymbol{\phi}(t)$ with a best fit linear model, and/or model discovery for generating a nonlinear dynamical system~\cite{kaheman2019learning}.

\subsection{Uncertainty quantification}

Model certification, credibility bounds, and other guarantees of performance are necessary for data-driven reduced order models in the aerospace industry.  Indeed, trustworthy machine learning is necessary for reduction to practice in almost any critical application area.
The mathematical framework of \emph{uncertainty quantification} (UQ) provides computational tools for evaluating probabilistic estimates of credibility and predictive capacity, and holds the key for bringing ML and AI into safety-critical domains.  
Without quantifying uncertainty in the model discovery procedure, one cannot provide estimates on the robustness and sensitivity of the models to observation error and model mismatch.
In practice, this limits the applicability of all of machine learning for applications where quantifying the credibility of predictions is critical, such as human transportation systems with stringent safety regulations. 

The mathematical architecture for UQ relies on a Bayesian perspective where predictions and quantification are given as probability distributions subject to a set of priors.  Data-driven discovery must then be equipped with physically meaningful priors.  It is also critical in developing a Bayesian scheme to separate dynamics from noise.  Mathematically, model parameters $\bm{\beta}$ will be conditioned on the data $\mathbf{Y}$ so that
\begin{equation}
  p(\bm{\beta} \mid \mathbf{Y}) \propto p(\mathbf{Y} \mid \mathbf{Z} (\bm{\beta})) \, p(\bm{\beta})
\end{equation}
where $\mathbf{Z}(\bm{\beta})$ denotes the predictions provided by the candidate models, $p(\mathbf{Y} \mid \mathbf{Z})$ is the observation likelihood given by the observation model, and $p(\bm{\beta})$ is the prior on the model coefficients.  A result of such UQ metrics is to have performance bounds and guarantees so that reduction to practice can be assessed and achieved.

\begin{figure}[t!]
\centering
\vspace{.1in}
    \begin{overpic}[width=\textwidth]{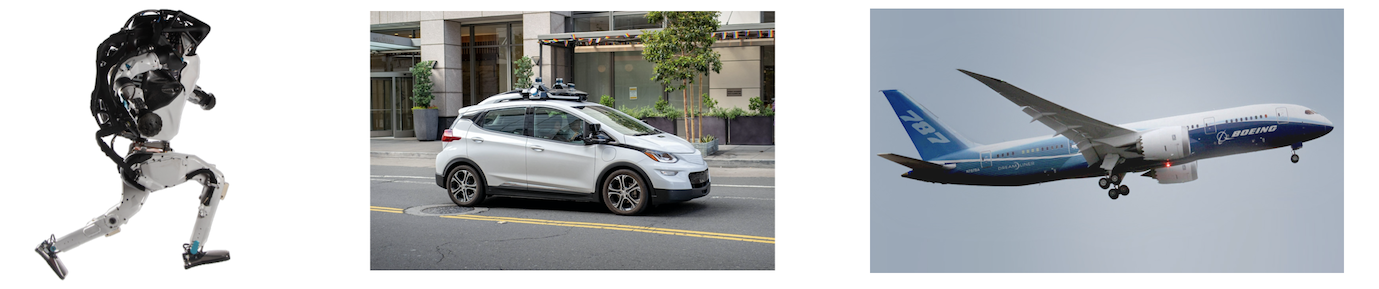}
\put(-1,22){\small (a) Robotics:  Physics-Informed}
\put(27,22){\small (b) Self-Driving Cars:  Physics-Free}
\put(63,22){\small (c) Autonomous Aircraft: Hybrid}
\end{overpic}
\vspace{-.3in}
    \caption{Robotics and self-driving cars, both broadly under the aegis of autonomy, are two leading data-intensive fields of the modern era.  (a) Robotics is largely physics-informed, with strong adherence to governing physical laws that enable robust control with data from high-fidelity sensors on the robot.  (b) In contrast, self-driving cars are largely physics-free, powered instead by computer vision algorithms  which have learned to characterize its environment and rules. (c) The future of aviation will require both key technologies.   }
    \label{Fig:Autonomy}
\end{figure}
\subsection{Autonomy and control}
Modern aerospace systems, including manufacturing and operations, will rely on advanced autonomy and precision control.  
To date, there has not been an emergent and well-established paradigm on how to most effectively use large-scale data for autonomous control systems.   Two modern grand challenge problems, robotics and self-driving cars, shown in Fig.~\ref{Fig:Autonomy}, exemplify two very different paradigms.  In the field of robotics, strong adherence to physics laws is imposed.  Thus a robot is strongly constrained by our classical physics-based models for its movement and balance.  However, critical use is made of high-fidelity sensors and Kalman filtering techniques, which jointly leverage models and data, in order to robustly control its sophisticated motions.   In contrast, self-driving cars are largely physics-free.  They are simply trained from exceptionally large data sets from sensors (vision, LIDAR, etc.), which attempt to integrate all possible scenarios the car may encounter in real-life driving scenarios.    Both paradigms for leveraging data, physics-informed and physics-free, have experienced tremendous success in the past decade.  They have also encountered fundamental limitations which must be overcome for the technologies to become commercially viable.  Self-driving cars have been empowered by deep learning algorithms, which are known to have significant shortcomings in extrapolation tasks.  Thus, some of the more spectacular failings of self-driving cars have come from situations which were not part of their training sets.  In contrast, our idealized physics-based models for complex and networked systems, such as robots, often fail to properly account for discrepancies between models and data, leading to control issues.   The aerospace industry ultimately will need to leverage both paradigms in order to fully exploit their large-scale data sets.

\begin{figure}[b]
    \vspace{-.7in}
    \centering
    \includegraphics[angle=90,width=.75\textwidth]{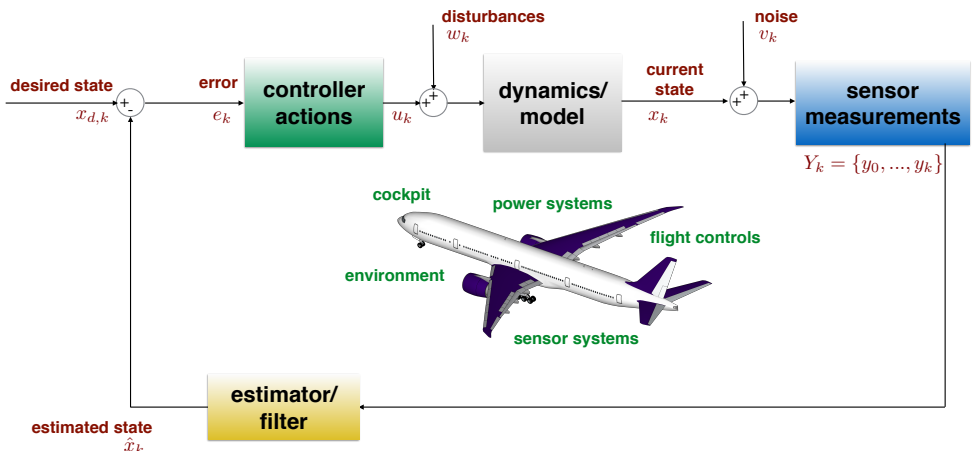}
    \vspace{-.95in}
    \caption{Standard classical control feedback loop.}
    \vspace{-.25in}
    \label{Fig:feedback}
\end{figure}

A traditional feedback system involves guidance (what state do we want a system to have), navigation (what is the current state of the system), and control (how will we affect the system to achieve the desired state), shown in Fig. \ref{Fig:feedback}.  
Key elements of autonomy and control that must be considered when leveraging of data science include:
\begin{itemize}\setlength\itemsep{-.065em}
    \item {\bf Performance criteria.} Performance criteria for certification are generally specified in either the frequency domain or the time domain.  In the time domain, the primary consideration is the transient response to a disturbance, relative to settling time, rise time, and overshoot.  In the frequency domain, steady state stability relative to harmonic forcing is considered  (bandwidth, stability to loop closure, gain margin, phase margin).  
    \item {\bf Safety criticality.}  Certification generally requires demonstration of a system meeting  design criteria to a certain probability of failure (e.g., ``five nines'' ($1 \times 10^{-5}$), ``seven nines'' ($1 \times 10^{-7}$) or ``nine nines'' ($1\times 10^{-9}$)).
    \item {\bf Human-in/on-the-loop.}  Effective control design must account for the interaction of the physical system with human pilots and passengers.  For example, in pilot-induced oscillation, the reaction time of a human pilot relative to system dynamics can result in the pilot's actions causing system oscillation to the point of instability.  Modeling human interaction with a physical system is exceptionally complex.
    \item {\bf Variable autonomy.} When humans and autonomy interact, the relative amount of human authority versus autonomous authority must be selected.  Typically systems are neither fully human operated nor fully autonomous.  The greatest challenges appear when the level of autonomy changes during active operation.  
\end{itemize}

\section{Aerospace Design}\label{Sec:Design}

Aircraft design starts with a set of requirements and ends with a production-ready vehicle design.  
Originally, the entire process  took only a handful engineers, from initial design, through multiple refinement and testing iterations, to final specifications.  
Modern aircraft design is much more complex and would be impossible without many advancements that have occurred in engineering, computers, applied mathematics, numerical methods, optimization, high-performance computing, geometric modeling, and more~\cite{raymer1989}. 

Several innovations enabled production of two seminal aircraft in the early 1930's.  The Boeing 247 and the DC-2 made commercial air travel viable.  Their all-metal design, a major factor, depended on advances in structural design, new manufacturing ideas, and development of new alloys.  The design process involved significantly more wind-tunnel and flight testing, as well as new results in potential theory that enabled broader, more accurate aerodynamics calculations--calculations performed on slide rule.  In fact, the aerospace industry has been a major driver in all of these fields due its level of design complexity~\cite{OswaldPhD, Oswald32}.

Advances in aerospace engineering, including jet propulsion, swept wings, radio navigation, super-sonic travel, computerized control systems, composite construction, {\it etc}, enabled by subsequent innovations in commercial aircraft design, have each required more accurate calculations, greater computational power, and increasing levels of interdisciplinary interactions.  Further advancements will require more innovations.  As data management tools are developed and adopted to handle the increasing amounts of data required for each iteration, the cumulative knowledge obtained remains diffuse and disparate.  And, while results in design exploration, surrogate modeling, and mixed-integer programming, have enabled and advanced multidisciplinary design, handling significantly more design parameters, as seems imperative, will require optimization based on necessarily sparse exploration of increasingly high-dimensional design space.  The aerospace industry is poised to both benefit from and drive advancements in data analytics, dimensionality reduction, data compression, and many more technologies broadly under the aegis of machine learning.   

Aircraft design has also expanded to cover the entire life cycle, including data management, manufacturability, technical oversight, evaluation criteria, operations and operability, maintenance, even disposal.  It involves many competing objectives and constraints, including safety, environmental impact, and ergonomics.  There is great potential from innovations in systems engineering, such as process control, robotics, flight scheduling, and flight path management.  Within the design process, planning and scheduling could benefit greatly from existing algorithms in decision making and control, with and without uncertainty, which have proved challenging to incorporate.

\subsection{Multidisciplinary design optimization}
Aircraft design is, in large part, a constrained, multi-objective optimization problem~\cite{Cramer94, Booker_et_al_98, Bowcutt_et_al_08, Henderson12, MDOsurveyJRRAM13, Bons_Martins_20}.  Constraints include airplane-level requirements, such as range or fuel capacity, as well as production and business constraints, such as manufacturing costs or product cash flow. 
 Design parameters may be discrete ({\it e.g.}, engine count), or continuous ({\it e.g.}, wing sweep); and the objective is some measure of expected profit or quantities presumed correlated with profit.  While the level of success depends on the efficiency and accuracy of the models, analysis tools, and objective functions, it also depends on the dimensionality of the design space and the ability to explore it thoroughly enough to identify a superior design with confidence.  Success also requires a robust design approach due to unavoidable uncertainties in predictions, algorithms, dynamics, {\it etc}.

Multidisciplinary design optimization (MDO) involves the use of numerical optimization in design when the constraints and objective(s) depend on two or more analysis disciplines, {\it e.g.}, aerodynamics, structures, propulsion, controls, cost (incl.\ research/development, design, manufacturing, operation, end-of-life), performance ({\it e.g.}, range, fuel burn), environmental impact ({\it e.g.}, noise, emissions), {\it etc}.  MDO can involve a diversity of parametric models that may include physical geometry, schematic layouts, or marketing mix.  Regardless of model type, effective MDO requires complete automation of every step from  parameter values to the objective function value for those parameters (or to the identification of the values as ``infeasible'').  This entails the models themselves, any processing required for their analysis, and post-analysis synergism.

\subsubsection{Design for manufacturability}

The increased use of composite materials, in particular the complexity of fabrication methods, brings manufacturability to the forefront of disciplines considered in multidisciplinary design optimization (MDO) studies. 
Reliable models are needed that can describe the output structure based on a variety of inputs, since the final structure of a composite material depends not only on the initial geometric design but also the material and fabrication method used.
Numerical modeling of fabrication methods, e.g. hand lay-up, automated fiber placement, and vacuum forming, requires deep knowledge of the physical processes that govern the final structure~\cite{IK17}.
Physics-informed machine learning may contribute to elevating first-principles-based models to the level of accuracy and efficiency required for an MDO study.

\subsection{Model-based engineering}

Model-based engineering (MBE) is an approach to product development and lifecycle management that focuses on using digital models and simulation to design, produce, maintain, and support products. 
Prior to the extensive use of digital artifacts across engineering domains, information and product specifications were transferred between engineering groups and product consumers through physical documents.
Digital models improve reliability and robustness of the entire product lifecycle by ensuring that designers, producers, and consumers have access to evolving representations of physical assets.
Techniques in data science are critical both in producing digital models that accurately capture asset behavior in the physical world and in standardizing digital model formats to ensure accessibility across the product lifecycle. 

\begin{figure}[t!]
\centering
\vspace{-.3in}
\includegraphics[width=.89\textwidth]{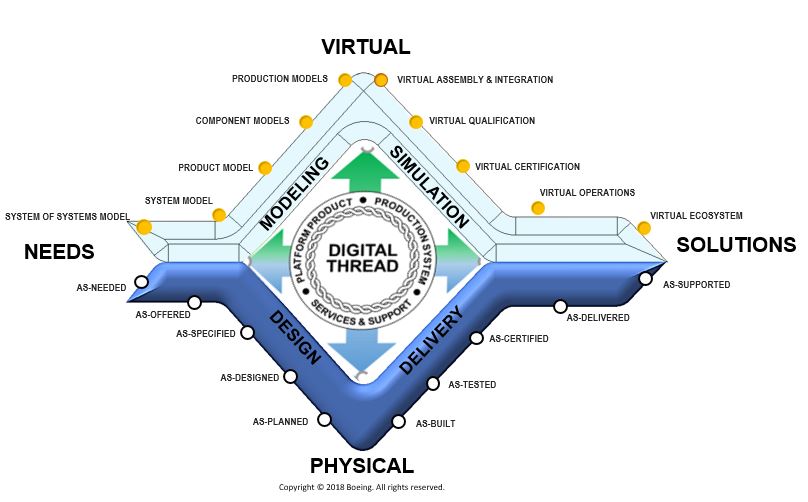}
\vspace{-.15in}
\caption{Model-based engineering design diamond.}
\vspace{-.15in}
\label{fig:design_diamond}
\end{figure}

\noindent\textbf{Design diamond.} The systems engineering design `V-model` depicts the major steps in the design and production of physical systems. 
The evolution of data generation in engineering processes as well as the technological development of robust digital models has led to a new view on the classical design V-model.
In Fig.~\ref{fig:design_diamond}, the standard systems engineering design V-model is mirrored to create a design diamond that incorporates digital counterparts of a product at all stages of its lifecycle.

\noindent\textbf{Digital thread.}
The digital thread is a digital communications framework connecting authoritative sources of information from producers to consumers in standard formats throughout the lifecycle of a process, product or system.
It is the connection layer that ties together the digital models and physical processes guaranteeing that necessary groups are working with a consistent set of evolving information to support the product.
One of the major challenges in implementing a digital thread is defining standard data formats that may interface with a wide variety of tools and processes.
However, the presence of a digital thread in MBE is essential for high fidelity digital twins.

\noindent\textbf{Digital twin.}
A digital twin is a virtual representation of the properties and behaviors of a specific instance of a physical system or process that enables prediction and optimization of performance and maintains synchronization with that physical system or process through its operational life.

The connection of a digital twin to its physical system is typically enabled through sensors gathering data in real-time during operations.
A complicated system, such as a commercial aircraft, may consist of hundreds of connected subsystems each producing a wide variety of signals detailing performance characteristics.
This leads to thousands of components producing diverse signals across varied communication channels, all in the course of real time operations~\cite{SHFTF18}.
Identification of an appropriate subset of these signals when modeling a particular system process is a critical challenge in constructing a reliable and useful digital twin.

\section{Aerospace Manufacturing}\label{Sec:Manufacturing}
Manufacturing is a highly complex and dynamic process, involving the coordination and merging of several elaborate and precisely times stages.  
In a modern manufacturing environment, tremendous volumes of data are being generated, stored, and analyzed to improve process quality, reliability, and efficiency.  
The data generated is inherently multi-modal, including hand-written reports from humans working alongside automation, high-fidelity data from metrology equipment, video feeds, supply chain logs, part catalogues, and results from in-process inspection, to name a few.  
There are several key opportunities to leverage machine learning and other data-intensive optimization techniques to improve manufacturing processes.  
Several areas of high-priority include: part standardization; automation and robotics; streamlined assembly, including reduced measurements, processing, and inspection, towards \emph{just-in-time} manufacturing; supply chain management; material design and fabrication; and non-destructive inspection.
In this section, we will explore several of these areas.  

\subsection{Advanced product quality planning}
Advanced product quality planning (APQP) is a methodology for producing physical systems that are guaranteed to meet target requirements.
It was developed in the late 20th century and has been used extensively throughout the automotive industry.
Techniques in data science will greatly impact almost every technique and framework contained in APQP.

Data science and machine learning technologies have significant potential to improve performance in statistical process control.
In aerospace manufacturing, this practice guarantees that parts are produced within a diverse set of specifications.
Historically, process control has been implemented for manufacturing processes by extensive testing in labs prior to production and then tuning these processes throughout production efforts. 
Without consistent data storage and formats, this may lead to the inability to transfer past process control efforts onto new platforms and processes. 
Data science and machine learning may assist in transferring past efforts by defining standard data formats and producing robust digital simulation models.

The increased use of automation in aerospace manufacturing has also enabled new opportunities for real time process monitoring. 
Manufacturing systems equipped with sensors gather real time process data, which can then be used to train machine learning based control models.
Trained on past production data, these models may determine when a process will move out of specifications prior to possible human measurement and detection. Furthermore, the features extracted using ML methodologies can help determine ideal  measurement and detection locations and can lead to significant reductions in labor and costs.

\subsection{Standardization}

The standardization of aircraft designs, manufacturing processes, parts and machines is an open industrial challenge. 
New designs are often airplane specific iterations that start the design process from scratch and ultimately offer no significant differences over prior versions.
However, current manufacturing paradigms do not exploit or streamline designs to reuse and retool existing versions, resulting in inefficiencies and stresses on supply chains, maintenance and the critical path of assembly. In aircraft design, some simple examples are brackets and fasteners. 
Brackets support wiring, hydraulics, systems, and fasteners for joining many types of structures and components. 
If data from prior designs can be extracted or harvested, then detailed comparisons can be performed and similarities can be identified resulting in a reduction of unique parts that are airplane specific.

Data mining and machine learning are poised to revolutionize industry paradigms for standardization. 
Currently, modern digital tools for designing, inventory and quality control produce a glut of data that often sits unused.
This data can be mined for dominant features to inform future decisions and streamline the design process.
To discover relevant, redundant patterns, feature selection criteria can be tailored to extract the desired topological, spatiotemporal, physical, and material properties. 
Recent strides have been made in interpretable machine learning models that directly incorporate sparsity-promoting, physical, temporal and topological constraints into optimization objectives.
The key features can be used to identify standard designs and reduce the number of unique designs or parts so that they can be shared across future designs and applications. 

\subsection{Automation}

The aircraft industry has driven the development of automated machines to fabricate detailed parts, including machining metals, composites fabrication, assembly of structure, painting, and inspection. 
These systems improve quality, reduce production cycle times and cost, and reduce repetitive injury to mechanics; however, they are expensive to develop, implement, and maintain. 
Due to the unique requirements of aircraft designs, materials, and tight tolerances, the decision to automate is complex and driven by specific manufacturing processes and applications. The development of successful automated systems starts with a focus on automating the process and understanding requirements. 
Thus, controlling the process is key to success. 
The tight tolerances of automation require monitoring and process control to ensure consistency, which in turn employs  many sensors. Sensor feedback can provide insight into where processes are stable and where variation occurs. 
They tell us the level of control needed in specific area, process repeatability, what and where to measure, and a host of other outcomes.

When control is present, algorithms for machine learning control can be used to disambiguate the relevant process dynamics from control inputs.
The tools of machine learning can also be used off-the-shelf on sensor data, not only to identify repetitive patterns and faults, but also pinpoint potential failures. Such failures can be predicted by studying the extent to which data deviate from dominant features, or by detecting outliers using robust feature extraction techniques on raw sensor data. The goal of robust feature extraction is to efficiently learn features and outliers simultaneously. One example is robust principal component analysis, which has previously demonstrated success in predictive aircraft shimming applications~\cite{Manohar2018jms} and is discussed in detail in Section~\ref{Sec:Shimming}. 

\subsection{Assembly}

Aircraft are manufactured in many pieces, including fuselage barrel sections, wings, stablizers, and fasteners, that are joined as sections or sub-assemblies and ultimately get integrated into the final product.  
Wings are built from sub-assemblies of fabricated wing skins, spars, flaps and ailerons; likewise, fuselages consist of multiple barrel sections that are assembled separately and then joined. 
Wings are joined to fuselages in a manufacturing position called ``€œwing-to-body join."
The alignment, positioning and joining of these assemblies require high precision tooling and part positioning systems.
In some cases, automated positioning tooling controlled by metrology systems are used to align and fit structure. In addition, even though large sections are made to very tight tolerances, the combined tolerances result in very small gaps that must be shimmed to meet engineering requirements. 
Parts are measured with metrology equipment, 3D models are generated to determine gaps between the mating parts, and automated machines are used to fabricate the shims that fill the gaps to ensure proper fit.

Throughout the process, enormous quantities of data are generated. 
Many sensors are used to provide feedback to control the automation, measure parts, and validate quality. 
The use of these data are expanding from building and assembling specific parts to being applied to predict future builds via machine learning and data analytics. Structural features mined from historical build data can be used to predict gaps in future builds, using only a targeted subset of measurements to infer the high fidelity structure. This {\em sparse sensing} methodology was successfully applied for the predictive shimming of new wing-to-body join builds (Section~\ref{Sec:Shimming}). Not only does sparse sensing bypass the heavy planning and processing required of high fidelity metrology, but it also identifies a set of key structural features that may be analyzed for diagnostics and defects.

\subsection{Materials}

\begin{figure}[t]

\centering
\includegraphics[width=.9\textwidth]{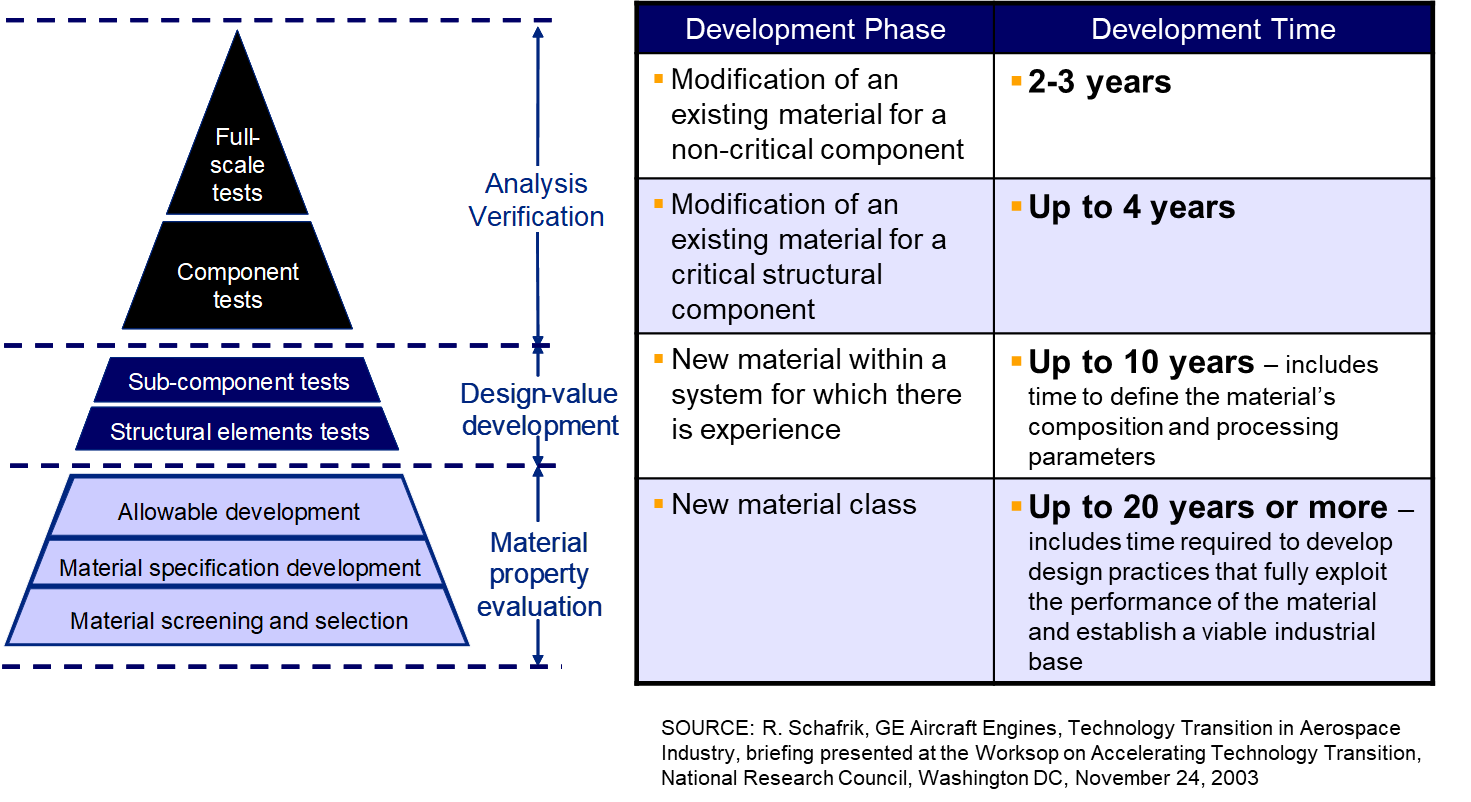}
\caption{Schematic materials building block diagram.  }

\label{fig:materials_building_block}
\end{figure}

Adoption of new materials is a major challenge in the aerospace industry, due to the amount of structural testing done to certify a material. 
The standard approach is the building block shown in Figure~\ref{fig:materials_building_block}.
This approach starts with a large amount of tests at small coupon levels, a much smaller number of tests at subcomponent scale, a handful of tests of subassemblies and one or two full-scale test articles. 
Moreover, additional validation is needed when the manufacturing process or process parameters change, because they can introduce different features in the part. 
There are two areas where machine learning can have significant impact: physical testing and materials characterization.

The use of data analytics and machine learning, in addition to physics-based models, can result in a significant reduction in testing. 
First, we might be able to correlate behavior at larger scales to a small number of fundamental material properties by analyzing data on existing materials.
That could already help filter out the most promising material candidates in the material screening and selection phase, without wasting time and money on extra characterization testing. 
In subsequent phases the sheer number of tests for new materials could be reduced by learning information from existing material systems. Secondly, in the absence of physical tests on new material parameters, physics-guided ML may be applied to process and structural simulations to correlate parameters to performance, thus enabling highly efficient exploration of material parameters. 
ML is also being used to design new materials~\cite{brunton2019methods}, including superalloys~\cite{conduit2017design,verpoort2018materials,conduit2018probabilistic,green2018quantum}.  

More recently there has been an increased focus on material characterization for fiber-reinforced composites process modeling to reduce the amount of physical trial and error.
Processing often happens when the material is uncured and at elevated temperature, requiring a different set of tests than is needed for the structural characterization.
For example, numerical models used to predict the quality of thermoplastic composite laminates~\cite{SU14, HSP13} require information on shear and bending behavior of the laminate, friction between layers (which changes when their relative orientation changes), friction between the composite and any contacting tooling material (which in turn could have some surface coating). 
Most of these properties change with temperature, and some also depend on the forming speed and pressure. Here, sparse regression techniques in ML can help discover fundamental laws and  relationships, resulting in significant reductions in the number of tests to be conducted, as well as the time and costs needed to introduce new materials.

\subsection{Composite fabrication}

Composite part fabrication can consist of many steps, some of which are more data rich than others. 
Examples of the most time consuming, defect-prone, or unpredictable processes are: material laydown, forming, uncured part handling, in-process inspection and autoclave curing. 
We will discuss some of these below.

Material laydown can be done manually, or by Automated Fiber Placement (AFP) or Automated Tape Laying (ATL). 
In AFP and ATL a robotic head places composite strips of material onto a tooling surface, building up each layer by making multiple passes to cover the surface. 
AFP is typically used for contoured parts, while ATL is most suitable for near-flat layup. Accuracy of the material placement is important for the quality and final strength of the part and requires in-process inspection. 
Traditionally the in-process inspection has been a labor and time intensive part of the process. 
Vision and thermal inspection systems~\cite{JGC20, CRH15} are starting to replace manual inspection to measure material placement and detect other defects. 
The data produced by these systems are prime candidates for data analytics and machine learning~\cite{SRBH20}, for diagnostics and defect detection.

\begin{figure}[t]

\centering
\includegraphics[width=.7\textwidth]{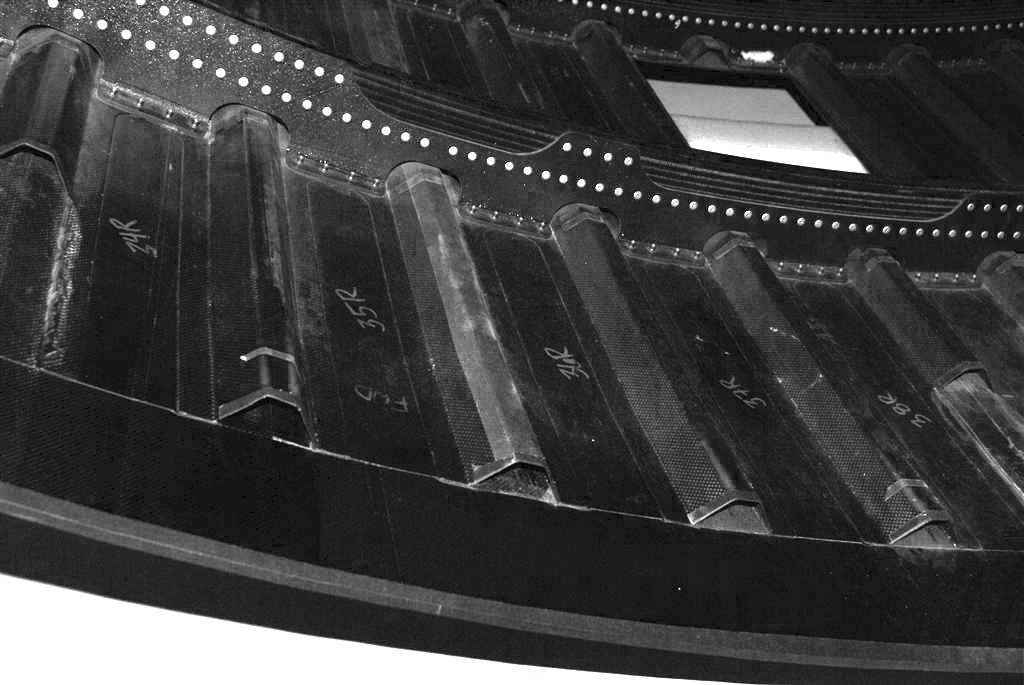}
\caption{Example of Hat-Shaped Stringers on a 787 Fuselage~\cite{B10}.}

\label{fig:hat_stringers}
\end{figure}

A significant number of composite parts are not laid up in their final contour, but instead laid up flat and then mechanically formed. 
Examples are stringers and spars, which are formed from a flat laminate to an L-profile, hat-profile (see Figure~\ref{fig:hat_stringers}), or C-profile, in a machine and subsequently placed in a tool so they can be combined with a second profile or a skin to form an assembly once they are cured. 
These profiles often have to follow the contour of a fuselage or wing skin, which besides the global aerodynamic shape can have relatively aggressive geometric transitions when the skin thickness changes. 
The profile forming is often done in a machine, if the global contour is benign, while aggressively curved stringers are manually laid up ply-by-ply.
Furthermore, placing the stringers in the tool is often done manually.  
The biggest risk with forming composite laminates is the generation of wrinkles when material is compressed locally to conform to its new shape. 
Small changes in pressure, material tack or composition of the layup cause variability in the presence, location and severity of the wrinkles. 
In addition, manual processes by definition have high variability, making it difficult to predict wrinkles for formed composites.
Therefore every part is inspected and structural integrity is verified for every occurrence, and if necessary, repairs are made to recover any loss in performance. 
Data analytics and machine learning have the potential to improve composite forming processes in three critical areas: (i) support the development of physics-based models by identifying features that are dominant in the formation of wrinkles, (ii) automate the detection and characterization of wrinkles in inspection data, and (iii) improve part quality by using vision-based systems to track humans and determine patterns in their actions that result in better quality parts, which could lead to better work instructions or the development of an automated system that mimics the best manual process.

\section{Aerospace Verification and Validation}\label{Sec:Testing}

The testing phase of aircraft development, which includes extensive validation, verification, and certification processes, is poised to leverage data-driven methodologies and the associated wealth of predictive analytics.  
Testing a new product involves verifying function and performance guarantees and validating that the system meets certification and government regulations. 
A critical goal in modern data-driven flight testing is to improve aircraft safety and robust operation, while reducing time and expense of testing programs. 
This goal will be enabled by a greater understanding of the fully integrated product behavior, its robustness, and nuances, resulting in more effective models and digital twins.  
This knowledge will inform and improve upstream design and manufacturing processes and downstream services, ultimately enabling faster, more flexible aircraft design and customization by streamlining verification and validation cycles.  
Testing occurs at the component level, subsystem level, and the integrated system level, as depicted in the design diamond in Fig.~\ref{fig:design_diamond}.  
Machine learning algorithms can be used to streamline the data collection and processing in each of these stages and to inform and refine earlier design and manufacturing stages.  
Other opportunities that are ripe for data-science enabled advances include: the extraction and visualization of patterns that may be inaccessible to human analysis; sensor optimization, robust processing, and anomaly detection; identifying and characterizing discrepancies between models and physical devices; and using active learning to streamline the number of experiments and data required to validate models and manage uncertainty.  
In what follows, we frame the evolution of testing mainly through the lens of commercial aviation products, whose key technology trends are empowered by data digitization, digital-twins, and physics-based modeling to account for dynamic behaviors. 

We first consider the type of data encountered in flight and lab tests. The amount of data generated by a flight test aircraft is diverse and vast, with upwards of $200,000$ multimodal sensor measurements during a single test.  The measurements are a mix of airplane generated digital production measurements and analog flight test instrumented sensors.  The multimodal sensors include strain gauges, pressure transducers, thermocouples, accelerometers, video, among many others.  Measurements are collected and stored asynchronously in time, with sampling rates varying from less than $1\,$Hz to upwards of $65\,$kHz.  Some measurement data is only collected when triggered or when a change is detected.  
All measurement data is stored for each flight test, generating gigabytes of data.  
Much of this data must be synthesized across sensor modalities and in time, both within a single flight test and across multiple tests.  
Similarly, high-rate asynchronous lab test data is collected in unique demanding environmental conditions, such as extreme vibration, icing conditions, sandblasting, extreme temperature, and high structural demands.  
These environments impose unique requirements on the sensors and can result in the sensors falling out of calibration.

The complexity of a typical commercial aircraft must also be considered.  There are approximately 2.3 million parts on the Boeing 787 with 70 miles of wire and 18 million lines of source code in the avionics and flight control systems alone.  Additionally, there are ten major systems interacting together in a complex, dynamic and rapidly changing environment.  As for constraints, critical system components must have a failure rate of less than $10^{-9}$ or $10^{-12}$, depending on the system, and backup components that seamlessly step in when required.  Flight testing is the final validation step for every design and performance requirement.  Certification is achieved when each component, subsystem, and integrated system process can be shown to have a proven level of performance, repeatability, and accuracy.  
Because aircraft are designed to have near zero tolerance for failures, these failures are intentionally triggered during testing to evaluate the resulting effects on function and performance.  
The rarity of failures in the high volume of flight test data presents an additional challenge to machine learning algorithms, which require a sufficient number of examples to build models for failure.  

\subsection{Digitization}

There is a significant opportunity to improve and automate verification and validation through data digitization, which will enable a wealth of downstream data analytics for pattern extraction and anomaly detection.   
To understand the benefits of a digitized workflow, it is helpful to understand the current testing landscape.  Existing operations are rampant with mundane manual touch points, requiring significant process, time, and resources to validate that there are no quality lapses.  
Current workflows rely on experienced engineers to synthesize upwards of 20 data sources to search for information to substantiate a report or troubleshoot a problem.
Systematic efforts to leverage data across flights and programs and to perform exploratory analyses are virtually non-existent. 

A digitized flight test workflow that leverages machine learning would enable the automation of tasks that are currently laborious, reactive, last minute, discrete, and manual.  
The result would be a flow of information that is effortlessly available to all test stakeholders and participants from the preliminary concept to the final test report.  Collection and aggregation of process metadata will enable visualization of core components, such as the test plan, test approvals, aircraft configuration, 
instrumentation configuration, and conformity status, with each component being updated as information becomes available.  
This centralized, real-time repository will provide engineers with the time and resources to process critical information, collaborate with peers, and engage with the test data as it is generated.  
 
There are several significant challenges that face efforts to digitize verification and validation workflows.  
Digitizing the workflow cannot compromise the integrity of the information or its ease of access.  
Any changes must improve upon the users' ability to confidently assess the quality, reliability, and traceability of the data.  
However, the benefits of making operations more efficient by leveraging machine learning and uncertainty quantification make digitation worth the effort. 
It is expected that digitization will result in sweeping benefits across the production cycle and in-service life cycles.

\subsection{Model-based validation}

Historically, testing requirements were developed in component or discipline-specific silos.  
These requirements were designed into test points as flight task cards to test specific configurations, or performance metrics.  
Critical or corner points of the performance envelope were identified as test conditions.  
Additional test points were determined by selecting evenly spaced points throughout the operating range of the component or aircraft.  
In conjunction with the test domain, the test procedures were aimed at isolating a small number of variables related to performance.  
The goal of test planning was to complete test conditions and test plans as efficiently as possible over the range of test parameters.  
It was typically the job of a test planner to consider readiness and dependencies of the test plans in order to identify the most efficient path through flight test.

Due to data capture and storage capacity limitations, data for flight tests were captured only for discrete single test conditions.  
Bandwidth and storage limitations prevented data from being used for real-time computing and recording.  
The phrase \textit{data on} was used as the marker to begin a test condition where data would begin recording.  
As such, data for test conditions was gathered and stored in discrete segmented test condition increments.  
Only a small percentage of all data was ever stored and analyzed on any given flight.  
Further, if you weren't \textit{data on} during a critical event in flight, either planned or unplanned, that data was lost without being recorded or analyzed.  
Engineers used paper strip charts that look similar to an EKG printout for real time plotting.  
If the strip chart was not set to plot the critical parameter during the test, there was no way to retrieve that data in flight.  Cross plots were often created by hand with paper and pencil.  Computers had limited computing capability and often only crucial calculations were computed in real-time.  
More intensive computing was performed by a mainframe computer after the test.

Technological innovations in computing and sensors have led to significant improvements in flight test data collection, storage, and visualization.  
Present day testing requirements are still largely developed in discipline-specific silos.  
Domain expertise has helped identify areas where tests can be modified to be more compatible or concurrent with other tests.  
Additionally, computational tools have been developed to aid human management and optimization of the flow of the test program.  
However, data science methods have yet to be used at significant scale to improve test optimization.  
Much of the improvement in present day testing has been the result of improved computing hardware.  
Storage has improved so all data parameters can be recorded and stored for the duration of an entire flight.  
Data analysis has moved from mainframe computers to laptops.  
There have been some improvements in algorithms and calculation methods especially in the areas of data fitting and filtering. 
Data visualization has seen the greatest improvements: paper has been replaced with electronic strip charts and plots, and the use of colors and  data triggers has allowed more data to be monitored and analyzed, with less cognitive overload. 
Although computational and visualization improvements have been significant, performance and evaluation metrics and algorithms have remained largely unchanged.  
Most data is still analyzed on a condition-by-condition basis, leaving large amounts of flight data unanalyzed.

Future testing will be heavily informed by the models which make up the digital twin.  
Safety-critical test conditions, where margins are thin, will be identified from models and simulations.  
Additional test conditions will be determined by a principled and data-informed selection, where physical tests will increase confidence in the model.  
Active learning~\cite{cohn1996active,settles2009active} may be used to streamline this testing, reducing the number of physical queries required to validate a model.  
Effective test planning will involve a multi-dimensional and multi-objective optimization to efficiently achieve convergence of the coupled, multi-scale and multi-physics models.  
Test plans will need to be flexible enough to support model exploration and refinement when the model diverges from the physical world; i.e., testing will need to inform a discrepancy model for the system.  
The goal of flight test will evolve from completing as many test conditions as possible to increasing certainty in the multitude of physical, functional, and logical models that make up the digital twin.  
Such future testing will require a more seamless synchronization with the digital twin to the physical entity it represents, providing behavioral or functional predictions, which are then informed by measurement data.  
Robust and computationally tractable reduced-order modeling approaches, along with data assimilation and uncertainty quantification, will be required to validate and explore models on the physical asset in real-time.  
Data from these models will then be used to update the digital twins.

\begin{figure}[t]
\begin{center}
\includegraphics[width=.8\textwidth]{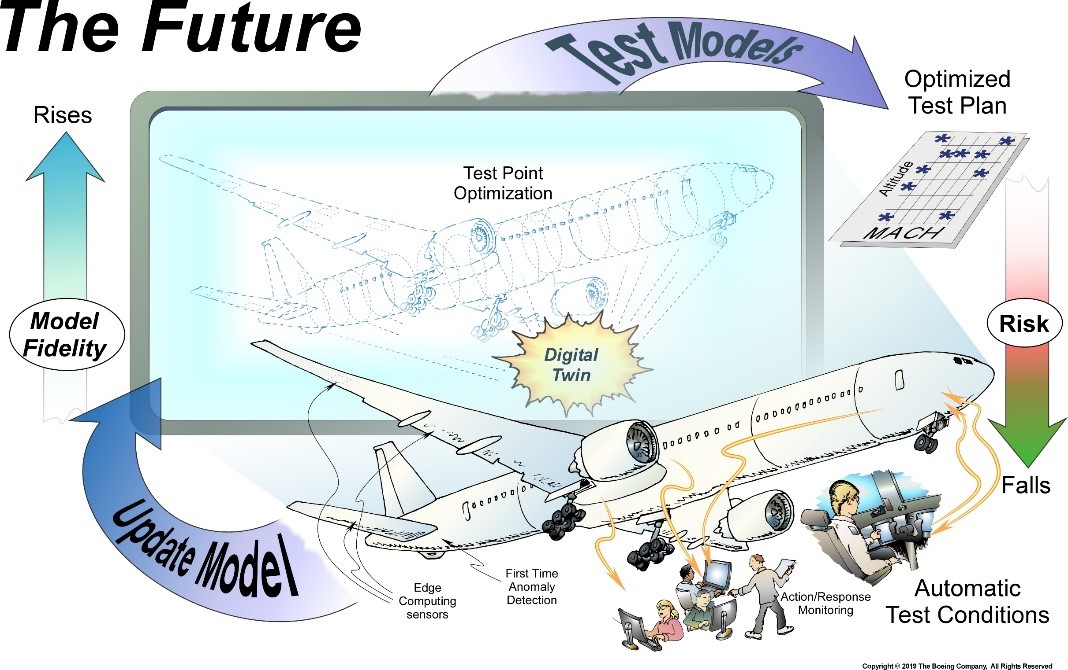}
\caption{Schematic overview of digital verification and validation processes, integrated with digital twin technology and optimized via machine learning. }\label{Fig:DigitalTwinCartoon}
\end{center}
\end{figure}

Flight test will likely continue to require discrete test conditions.  
However, an increased emphasis should be placed on analyzing data beyond what is required for the test condition. 
This holistic approach differs significantly from the past approach of only capturing the ``data on'' condition data to now making 100\% of the data available for analysis. 
Modern techniques in machine learning will facilitate this analysis of high-dimensional, multi-modal data, potentially extracting patterns and correlations that would have been inaccessible with previous approaches.  
For this model and data revolution to be successful, additional data from disparate sources will be required beyond what is needed to validate the digital twin.  
For example, meta data will link these data, including information about what parts were installed (configuration), what airport(s) were involved, and what were the qualitative characteristics of the aircraft.  
Combining this meta-data with the digital twin will be crucial to flight test of the future.

\subsection{Leveraging emerging technologies and algorithms}

There are several other opportunities for enhanced flight test capabilities that are being enabled by rapidly advancing technologies and algorithms.  
As aerospace products become more advanced, it becomes increasingly difficult for human experts to manage the interactions between systems when troubleshooting a problem or designing a test with multiple failure conditions.  
However, machine learning and data visualization algorithms are enabling the analysis of vast, high-dimensional data sets, resulting in the identification of patterns and correlations that are intractable to human analysis.  
Identifying potential interactions, and ultimately, failure modes, is a combinatorially complex task; however, artificial intelligence systems of the future may seamlessly integrate decades of past test data to make informed decisions about likely complications.  
This advanced diagnosis and problem solving capability will only be possible with system wide digital twin and digital thread efforts.  
In addition, considerable test time may be saved by automatically identifying sensor failure and detecting anomalous data, such as calibration drifts.  
Improved sensor fusion and filtering algorithms, based on physics-informed machine learning models of the system and its components, will enable real-time diagnoses with confidence intervals.  
Sparse sensor optimization, or identifying which sensors to listen to for a given task, will also reduce the computational burden, enabling real-time decisions and updates to the flight plan, instead of post-mortem analysis.  

\section{Aerospace Services}\label{Sec:Services}
In-service operations are the support provided to customers or by the customer operations team to support an operational fleet.  From the initial delivery to long-term maintenance, there exists significant opportunities for leveraging data to improve and optimize maintenance and airside support.  A major source of customer expense arises from aircraft damage, injury, delays to airside support, and unscheduled maintenance.  The positive business impacts of autonomous airside operations and not incurring unscheduled maintenance can be met with emerging sensor and algorithmic innovations. 

\subsection{Autonomous airside support}

Immediately upon arrival at an airport, preparations begin for receiving the airplane and preparing it for its departure.  Critical tasks include (i) accurately positioning the aircraft in and out of the gate, (ii) removing passengers, cargo, and waste, (iii)  reloading the aircraft with passengers, luggage, potable water, etc., (iv) servicing any aircraft fluids (oil, washer fluid, fuel) or maintenance items, (v) investigating any reported inbound issues, and (vi) performing a visual pre-inspection check in the interior and exterior of the aircraft.  Many of these tasks can be automated with modern technology, either through robotic interactions, autonomous vehicles, or visual inspection using current computer vision algorithms.   Importantly, modern IoT technology can integrate the diverse data streams in order to make an informed and accurate decision about the airside support tasks that require human intervention.  Such data integration can provide significant savings, which allows human resources to be used in targeted and efficient ways.  Importantly, even in this \textit{fully} autonomous system, humans and autonomy will need to interface in a safe, efficient, and effective manner.   Consideration of robotic failure conditions and human activity awareness will be required.

\begin{figure}[t]
\centering
\includegraphics[width=.9\textwidth]{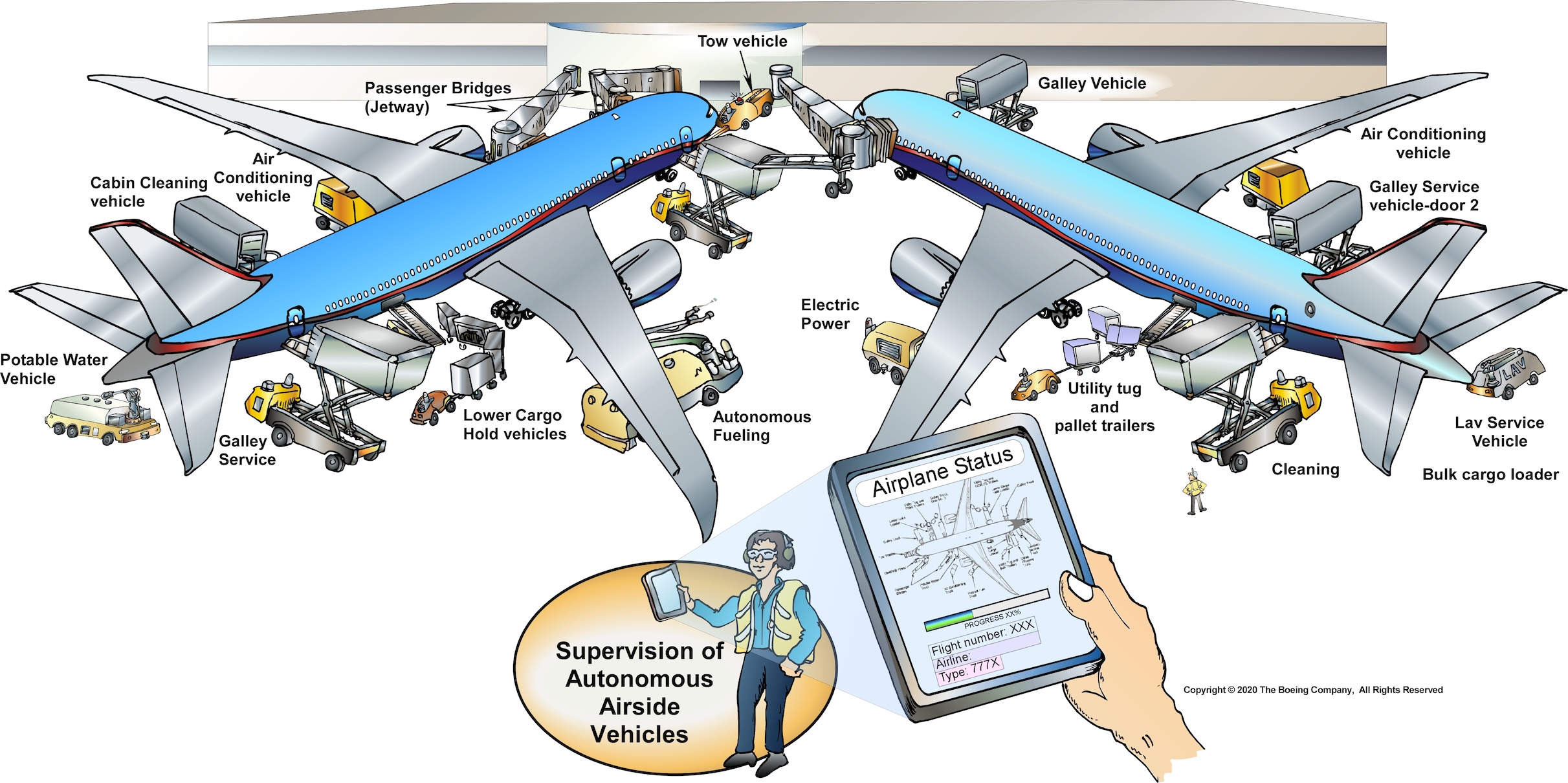}
\caption{Illustration of many potentially data-intensive aspects of ground service.}\label{fig:GroundService}
\end{figure}

\subsection{Eliminating unscheduled maintenance}
Unscheduled maintenance causes costly delays, wreaks havoc on a carefully optimized schedule, and inconveniences passengers, damaging the reputation of the airline carrier.  Customers must balance the cost and time of preventative maintenance with the cost of unscheduled maintenance.  Eliminating unscheduled maintenance would transform the industry, and this may be enabled by an expert aircraft system that identifies and communicates part anomalies and self-diagnoses wear, low fluids, software glitches, etc. 
This technical advance would require customer support centers to be able to download and process massive near-real-time data, run trend analysis, and make automated decisions in real-time.  
 Current prescriptive models indicate that a particular part will fail within the next three flights with an 80\% accuracy; while this information seems valuable, it is only marginally useful to improve operations.  Increased value would be gained by understanding why the part is failing, or explaining the exact maintenance actions that need to be taken to rectify the situation. Knowing the right maintenance actions, for example, greasing the right bearing, updating software or replacing the whole part, would have a significant impact on the efficiency and efficacy of maintaining the fleet. Being able to detect that a failure is about to happen, or has happened, is good.  Being able to identify why the part is failing, and enabling a recommender system to assess the best course of action, is even better.  Often nuances in how the operator uses the system violates the assumptions that were made when determining the maintenance schedules which leads to early part failures.  Leveraging part data would enable root cause identification and inform the design centers to enable future designs to be better aligned with in-service operations.  
 
\subsection{Closing the loop with design, manufacturing, and testing}

When the digital thread connects data from aircraft design, manufacturing, and testing, it will be possible to leverage service data to improve each of these stages, and vice versa.  
The lifetime of an aircraft, from concept designs to retirement, span decades, and the aging processes of new materials and manufacturing processes must be incorporated into the digital twin.  
These data-informed models will enable adjustments and refinements to the design and manufacturing procedures to improve the performance of future aircraft.  
Similarly, a more holistic digital twin, including models for aging and degradation, will be useful for maintaining and services fleets more effectively.  
These models will facilitate more accurate sensor filtering and data assimilation, as well as downstream tasks of anomaly diagnosis and detection.

\section{Case Study: Predictive Assembly and Shimming}\label{Sec:Shimming}
Aircraft are built to exceedingly high tolerances, with components sourced from around the globe.  
Even when parts are manufactured to specification, there may be significant gaps between structural components upon assembly due to manufacturing tolerances adding up across large structures.  
One of the most time-consuming and expensive efforts in part-to-part assembly is the shimming required to bring an aircraft into the engineering nominal shape.  
A modern aircraft may require on the order of thousands of custom shims to fill gaps between structural components in the airframe. 
These shims, whether liquid or solid, are necessary to eliminate gaps, maintain structural performance, and minimize pull-down forces required to bring the aircraft into engineering nominal configuration for peak aerodynamic efficiency.  

Historically, parts have been dry-fit, gaps measured manually, and custom shims manufactured and inserted, often involving disassembly and reassembly.  
Recent advancements in 3-D scanning have enabled their use for surface measurement prior to assembly, known as {\it predictive shimming}~\cite{marsh2008laser,jamshidi2010manufacturing,muelaner2010design,marsh2010method,muelaner2011measurement,chouvion2011interface,muelaner2011integrated,muelaner2013achieving,boyl2014digitally,vasquez2014systems,valenzuela2015systems,boyl2016methods,antolin2016end}.  
Gap filling is a time-consuming process, involving either expensive by-hand inspection or computations on vast measurement data from increasingly sophisticated metrology equipment.  
In either case, this amounts to significant delays in production, with much of the time being spent in the critical path of the aircraft assembly.  

\begin{figure}[b!]
\begin{center}
\vspace{-.1in}
\includegraphics[width=.95\textwidth]{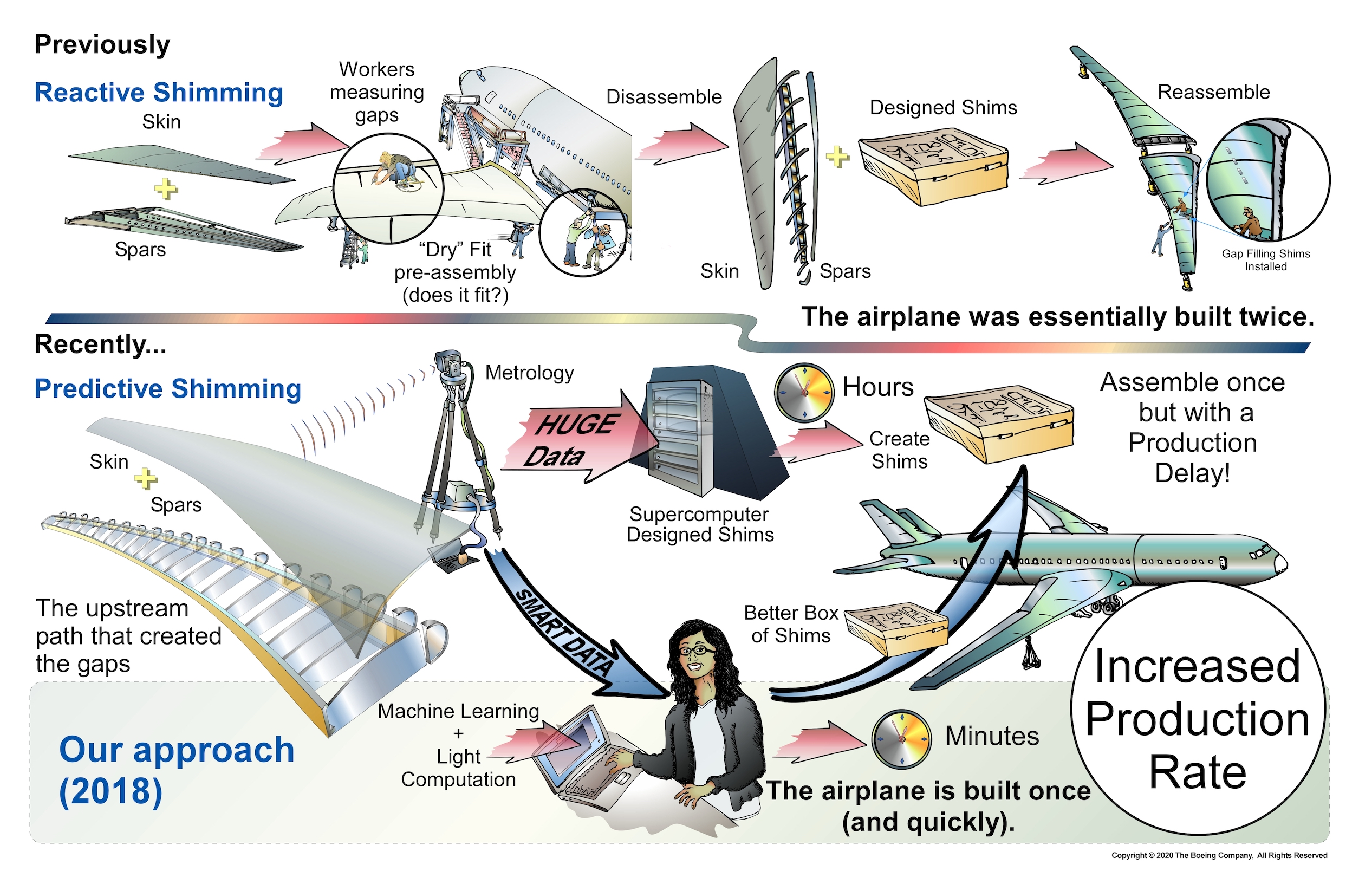}
\end{center}
\vspace{-.275in}
\caption{Cartoon illustrating recent progress in predictive shimming in the last decade.}\label{Fig:ShimmingCartoon}
\vspace{-.15in}
\end{figure}

In this case study, we present a recent strategy for predictive shimming~\cite{Manohar2018jms}, based on machine learning and sparse sensing to learn gap distributions from historical data and then design optimized sparse sensing strategies to streamline the collection and processing of data; see Fig.~\ref{Fig:ShimmingCartoon}.  
This new approach is based on the assumption that patterns exist in shim distributions across aircraft, and that these patterns may be mined from historical production data and used to reduce the burden of data collection and processing in future aircraft.  
Specifically, robust principal component analysis (RPCA)~\cite{rpca} from Sec.~\ref{Sec:Methods:RPCA} is used to extract low-dimensional patterns in the gap measurements while rejecting outliers.  
RPCA is based on the computationally efficient singular value decomposition (SVD)~\cite{Golub1965siamb,Brunton2019book}, and yields the most correlated spatial structures in the aircraft measurements, identifying areas of high variance across different aircraft.  
Next, optimized sparse sensors~\cite{Brunton2016siap,Manohar2017csm,Manohar2016jfs} are obtained that are most informative about the dimensions of a new aircraft in these low-dimensional principal components.  
The success of the proposed approach, {known within Boeing as}{} PIXel Identification Despite Uncertainty in Sensor Technology (PIXI-DUST), is demonstrated on historical production data from 54 representative Boeing 
 commercial aircraft.  
This algorithm successfully predicts $99\%$ of the shim gaps within the desired measurement tolerance using around $3\%$ of the laser scan points that are typically required; all results are rigorously cross-validated. 

This approach to predictive shimming combines robust dimensionality reduction and sparse sensor optimization algorithms to dramatically reduce the number of measurements required to shim a modern aircraft.  
In particular, RPCA from Sec.~\ref{Sec:Methods:RPCA} is used to extract coherent patterns from historical aircraft production data.  
Thus, RPCA is used to develop low-dimensional representations for high-dimensional aircraft metrology data (e.g., laser scans or point cloud measurements). 
Shim scan data is collected across multiple aircraft, either leveraging historical data, or collecting data in a streaming fashion as each new aircraft is assembled. 
The shim measurements for a region of interest are flattened into column vectors $\bx_k\in\mathbb{R}^n$, where $n$ corresponds to the number of measurements and $k$ refers to the aircraft line number. 
These vectors are then stacked as columns of a matrix $\bX = \begin{bmatrix} \bx_1 & \bx_2 & \cdots & \bx_m\end{bmatrix},$
where the total number of aircraft $m$ is assumed to be significantly smaller than the number of measurements per aircraft, i.e. $m\ll n$. 

Because of tight manufacturing tolerances and a high degree of reproducibility from part to part, it is assumed that the matrix $\bX$ possesses low-rank structure. 
As described above, there may be sparse outliers that will corrupt these coherent features, motivating the use of RPCA to extract the dominant features.  
Next, sparse sensors are identified which maximally inform the patterns in future aircraft using sparse optimization techniques from Sec.~\ref{Sec:Sensors}.  
The goal is to identify a small number of key locations that, if measured, will enable the shim geometry to be predicted at all other locations; this is possible because of the low-dimensional structure extracted through RPCA. 
Only measuring at these few locations will dramatically reduce measurement and computational times, improving efficiency.    

\begin{figure}[b!]
\centering
\begin{subfigure}[t]{.375\textwidth}
\includegraphics[width=1\textwidth]{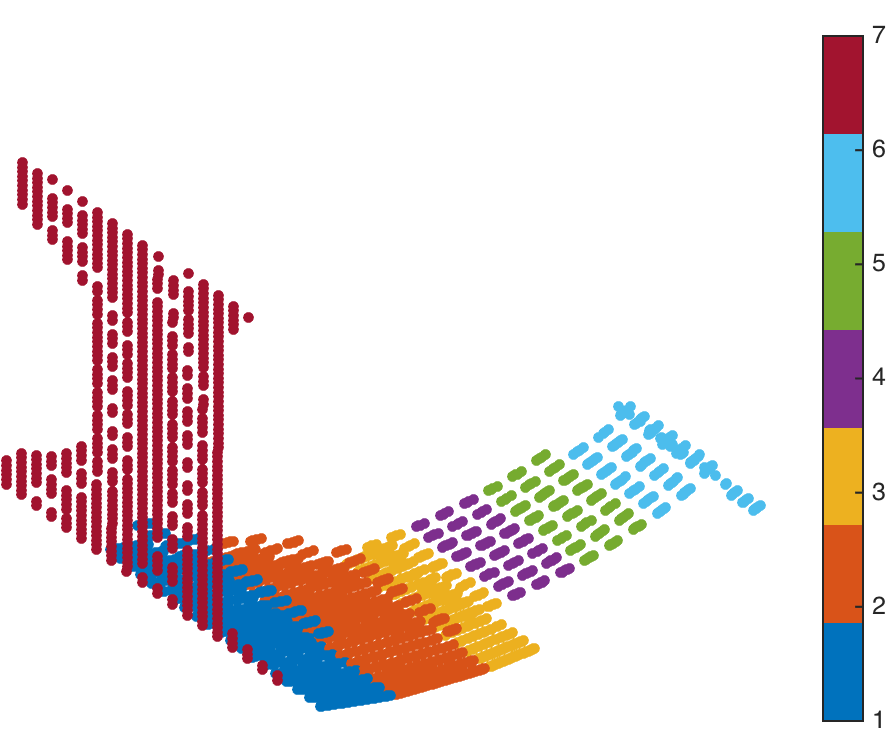}
\caption{High-fidelity scan locations}
\end{subfigure}
\quad\quad
\begin{subfigure}[t]{.375\textwidth}
\includegraphics[width=1\textwidth]{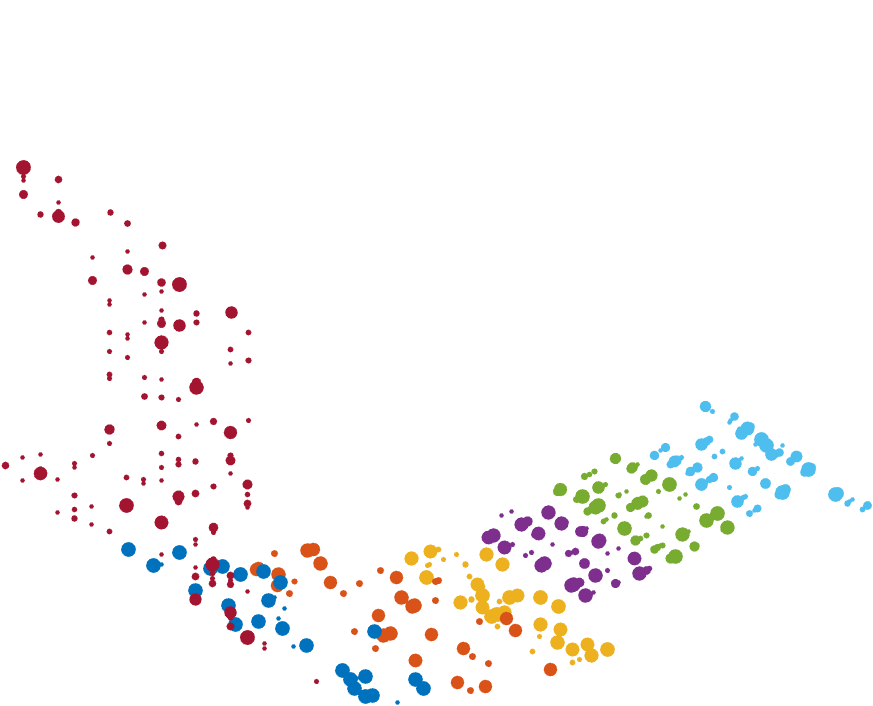}
\caption{Optimal measurement sensor ensembles}
\end{subfigure}
\caption{{\bf Gap measurements}. {This figure plots high-fidelity gap measurement locations segmented into 7 canonical shim regions (left). Ensembles of measurement locations (sensors) selected by our method are pictured on the right. Sensors are sized by the number of times they are selected by our method across 53 different gap measurement training sets. They are more concentrated around edges and corners, which is where we expect more systemic gap variation. }{Shim boundaries for segmented prediction (left) and sensor ensembles across 53 instances of training (right), sized by the number of times they appear in training. \textit{Reproduced with permission from~\cite{Manohar2018jms}.}} \label{fig:segmented_pred}}
\vspace{-.15in}
\end{figure}

We demonstrate the PIXI-DUST architecture on production data from a challenging part-to-part assembly on a Boeing aircraft.  
This data set consists of $10076$ laser gap measurements of the part assembly for {54 different production instances of the same aircraft type. Measurement locations are aligned between the instances, making the data amenable to SVD.}  
We build a low-order model of the shim distribution using RPCA and then design optimized measurement locations based on these data-driven features.  
We train the model on $53$ aircraft and then validate on the remaining aircraft; this process is repeated for all $54$ possible training/test partitions to cross-validate the results.  
Thus, a data matrix $\bX\in\mathbb{R}^{10076\times 53}$ of training data is constructed, in which each column contains all of the shim gaps for one aircraft, and each row contains the measured gap values at one specific location for all aircraft in the training set. 

\begin{figure}[t]
\begin{center}
\begin{overpic}[width=\textwidth]{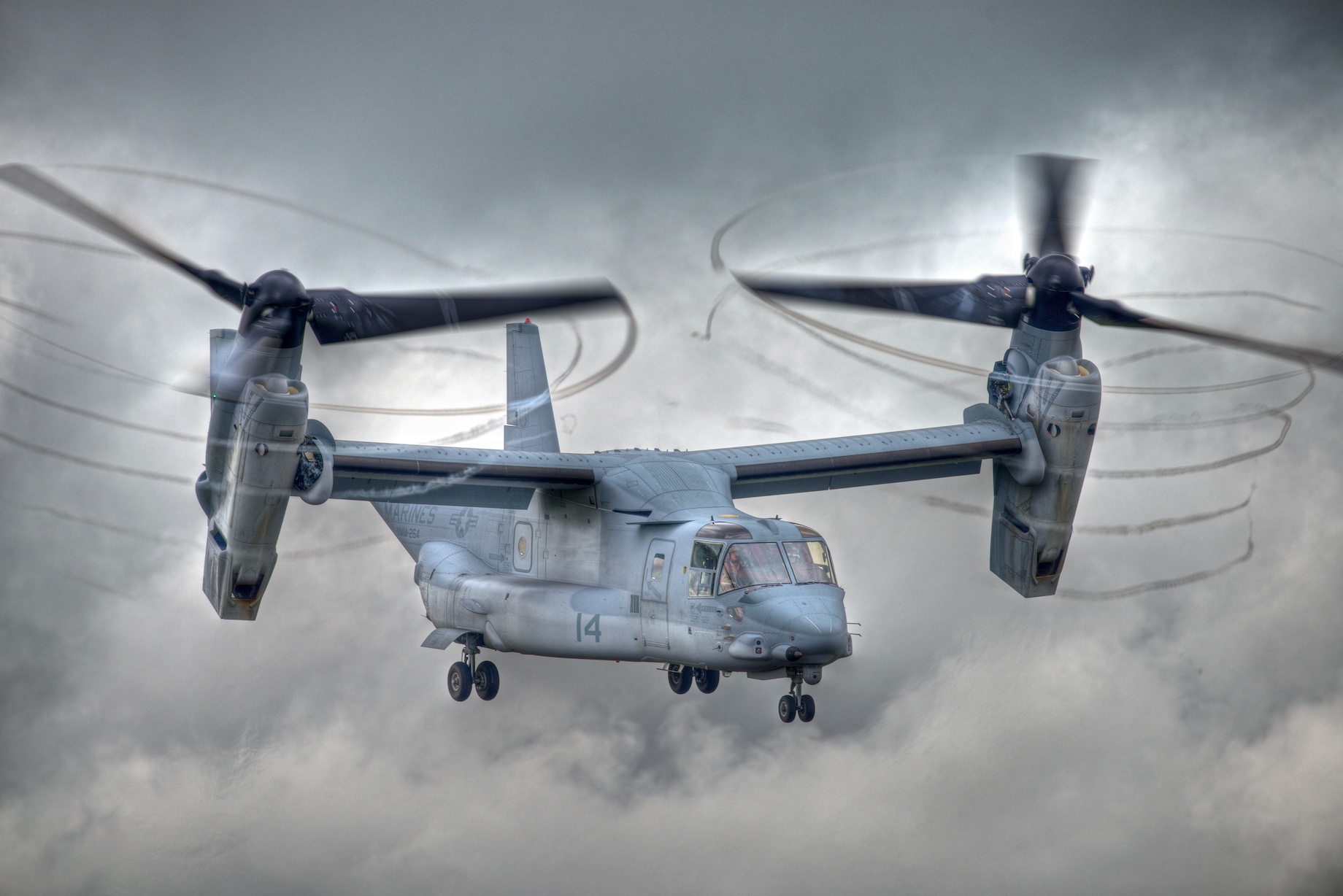}
\put(4,38){Shim 1}
\put(28.5,38){Shim 2}
\put(53,38){Shim 3}
\put(77,38){Shim 4}
\put(4,18){Shim 5}
\put(28.5,18){Shim 6}
\put(53,18){Shim 7}
\put(5,-2){\small  $\log_{10}|x_i-\hat{x}_i|$, Error (in.)}
\end{overpic}
\vspace{.05in}
	\caption{{\bf Absolute error distribution for optimal (blue) vs. random (orange) sensors.} The above are histograms of pointwise absolute error across all validation tests using reconstruction from optimal and an equal number of {\it random} sensors {(which are randomly selected from available measurement locations shown in Figure~\ref{fig:segmented_pred})}{}. The red line represents the desired measurement tolerance, and the legend indicates the percentage of points that fall within this error tolerance. The histograms indicate that optimal sensors predict nearly twice as well as random sensors. \textit{Reproduced with permission from~\cite{Manohar2018jms}.}}\label{fig:shim_error_hist}
\end{center}
\end{figure}
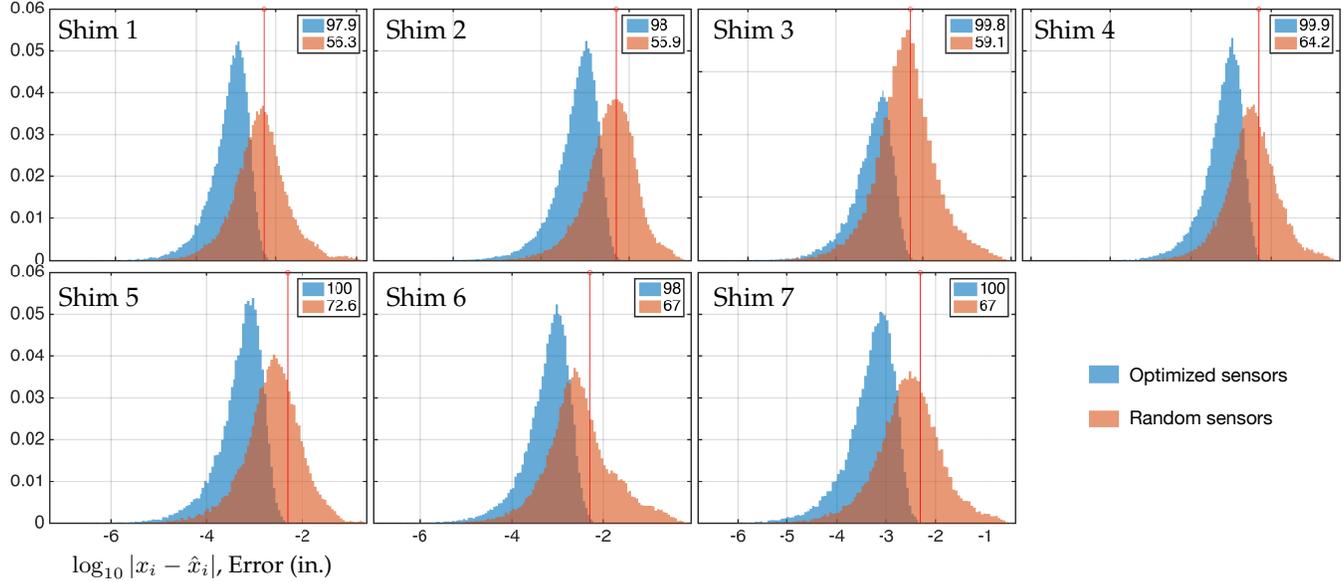

Figure~\ref{fig:segmented_pred} displays the seven separately manufactured shim segments, as well as the sensor ensembles for each shim; the error distributions are given in Fig.~\ref{fig:shim_error_hist}. Prediction results are shown in Table~\ref{Tab:segmented}. Prediction accuracy is vastly improved, and 96-99\% of the shim gap locations are predicted to within the desired machining tolerance. Furthermore, we note that the rates of optimal measurements $r$ vary from anywhere between 2\% to 6\% of all points within the shim, which indicates that some shims are higher-dimensional and require more features (hence, sensors) to be fully characterized. This is also reflected in the sensor ensembles in Fig.~\ref{fig:segmented_pred}.

\begin{table*}[t]
	\centering
	\scalebox{1}{
		\begin{tabular}{l c c c c c c c} 
			\hline 			\hline
            \multicolumn{1}{l}{\bf Shim No.}
			& \multicolumn{1}{c}{\bf 1}
			& \multicolumn{1}{c}{\bf 2}
			& \multicolumn{1}{c}{\bf 3}
			& \multicolumn{1}{c}{\bf 4}
			& \multicolumn{1}{c}{\bf 5}            
			& \multicolumn{1}{c}{\bf 6}            
			& \multicolumn{1}{c}{\bf 7}
			\\
			\hline
			
			\multirow{1}{*}{\rotatebox[origin=c]{0}{\parbox{4.5cm}{Percent Accurate}}} 
			& 97.90		& 98.05	&  99.82  &  99.94 & 99.99  &  99.03 & 99.97 \\ 					
			\multirow{1}{*}{\rotatebox[origin=c]{0}{\parbox{5.25cm}{Optimal sensors (avg)}}} 
			& 26     &  26	&  25   & 26 &  25  & 26 & 25   \\ 
			\multirow{1}{*}{\rotatebox[origin=c]{0}{\parbox{4.5cm}{Total points}}} 
			& 1003     &  1116	&  453   & 692 & 709   & 768 &  664   \\ 								%
			\hline \hline
		\end{tabular}
	}
	\caption{Segmented prediction results show vastly improved prediction accuracies, with 97-99\% of gaps predicted to within the desired $0.005$ inch measurement tolerance.}
	\label{Tab:segmented}
\end{table*}

This case study demonstrates the ability of data-driven sensor optimization to dramatically reduce the number of measurements required to accurately predict shim gaps in aircraft assembly.  
Sparse sensor optimization was combined with robust feature extraction and applied to historical Boeing production data from a representative aircraft.  
With around $3\%$ of the original laser scan measurements, this learning algorithm is able to predict the vast majority of gap values to within the desired measurement tolerance, resulting in accurate shim prediction.  
These optimized measurements exhibit excellent cross-validated performance and may inform targeted, localized laser scans in the future. 
Reducing the burden of data acquisition and downstream computations has tremendous potential to improve the efficiency of predictive shimming applications, which are ubiquitous and often take place in the critical path of aircraft assembly.  
Thus, streamlining this process may result in billions of dollars of savings. 

\section{Case Study: V-22 Osprey}
The V-22 is a multirole tiltrotor combat aircraft, designed to take off and land like a helicopter, but with the range and speed of a fixed-wing aircraft.  
These impressive capabilities came with considerable engineering challenges, including the mechanics of the tiltrotor, the aerodynamics, and the control systems.  
The development and test of the V-22 were mired in delays, cost overruns, and safety mishaps.  
The purpose of this case study is to identify how the use of data science and a digital twin could have benefited the design and testing of a technologically advanced new aircraft, such as the V-22.  
The intent of this retrospective is not to critique the V-22, which is by all accounts an engineering marvel, or the design and testing program; many of the modern technologies discussed here did not exist or were in their infancy at the time.  
Instead, we present a summary, based on lessons learned from the V-22 program, where modern digital twin technology would have been beneficial to safety and facilitated an on-schedule design.  

\begin{figure}[t!]
\begin{center}
\includegraphics[width=.85\textwidth]{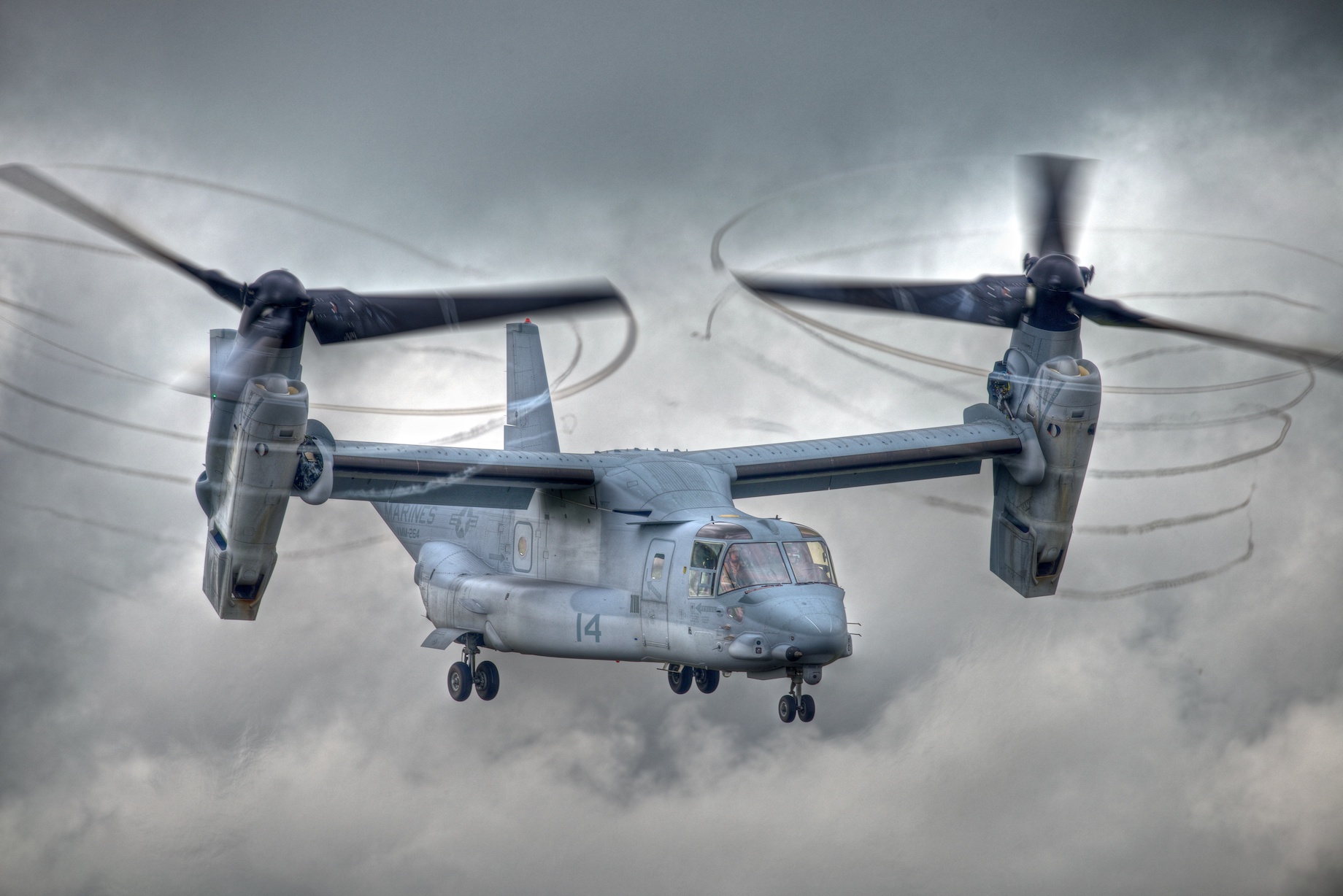}
\vspace{-.1in}
\caption{V-22 Osprey. \textit{Image reproduced from https://en.wikipedia.org/wiki/File:V22-Osprey.jpg}.}
\vspace{-.1in}
\end{center}
\end{figure}

A 2001 review panel on the V-22 recommended the program ``Extend high-rate-of-descent testing, formation flying (and other deferred flight tests as appropriate) to sufficiently define and understand the high-risk portion of the flight envelope under all appropriate flight conditions."~\cite{johnson2005model} 
The panel went further and recommended that the results of the high risk, high rate-of-descent testing be used to update operating limitations as well as the flight simulator used for crew training.  
The call for additional high rate-of-descent testing came as a result of a fatal accident in April 2000 involving both high-rate-of-descent and formation flying.  
The cause of the accident was attributed to an aerodynamic condition known as a vortex ring state (VRS), which is an unsteady aerodynamic condition that occurs for rotorcraft operating at low forward speed and high rate-of-descent.  
In this condition, the rotorcraft does not fly away from the rotor wake, resulting in highly unsteady airflow.  
For a helicopter, this often results in a high rate-of-descent, and for a tiltrotor the result is typically roll-off.  
Both situations are perilous, especially in close proximity to the ground.  
VRS still remains a significant area of research.  

The goal of a digital twin is to bridge the gap between the physical world and the virtual world.  
Two aspects of a digital twin that differentiate it from traditional modeling are: 1) continuously learning and updating internal models with real-world data, and 2) integration of data from one or more multi-physics models across a system.  
For a digital twin to continuously learn, it needs to be tested, updated, and re-tested.  
In the case of the V-22, a digital twin may begin with a model of VRS such as the one presented in~\cite{johnson2005model}, which is merged with the model and flight test data for the specific aircraft.  
Discrepancy modeling can be used as a data-driven method to explain and update model divergence. 
On at least two occasions V-22 pilots experienced significant uncommanded roll events during formation flying~\cite{V22Report}.  
As a result of this, and the aforementioned accident, the V-22 kicked off an extensive formation flying and high-rate-of-descent flight test campaign.  
Combining a VRS model with formation flying models would be an ideal opportunity to integrate multi-physics models across a system.  
The goal of integration as data becomes available is to work towards model convergence, or the bridging of the physical and virtual such that the digital model accurately represents the physical environment.  
The integration of unsteady, nonlinear, multi-physics models is no easy task.  
Implementing a real-time reduced-ordered model may help discover and explore areas of significant model discrepancies in flight test.  
The use of physics informed learning and custom regularization in the machine learning optimizations  may aid in maintaining physical constraints as the two models are merged to form a digital twin.  
With that enhanced model in hand, there is benefit in learning from a high number of simulations.  
Flight testing could be reduced to the most critical test conditions: safety, mission critical, and model confirmation.  
Certainly, this simple case study is not without conjecture.  However, the tools presented in this paper will form the basis for data-driven hypotheses that serve as the roadmap for future aerospace engineering possibilities.    

\section{Case Study: Urban Air Mobility}
\begin{figure}[b]
\vspace{-.35in}
\begin{center}
\includegraphics[width=.95\textwidth]{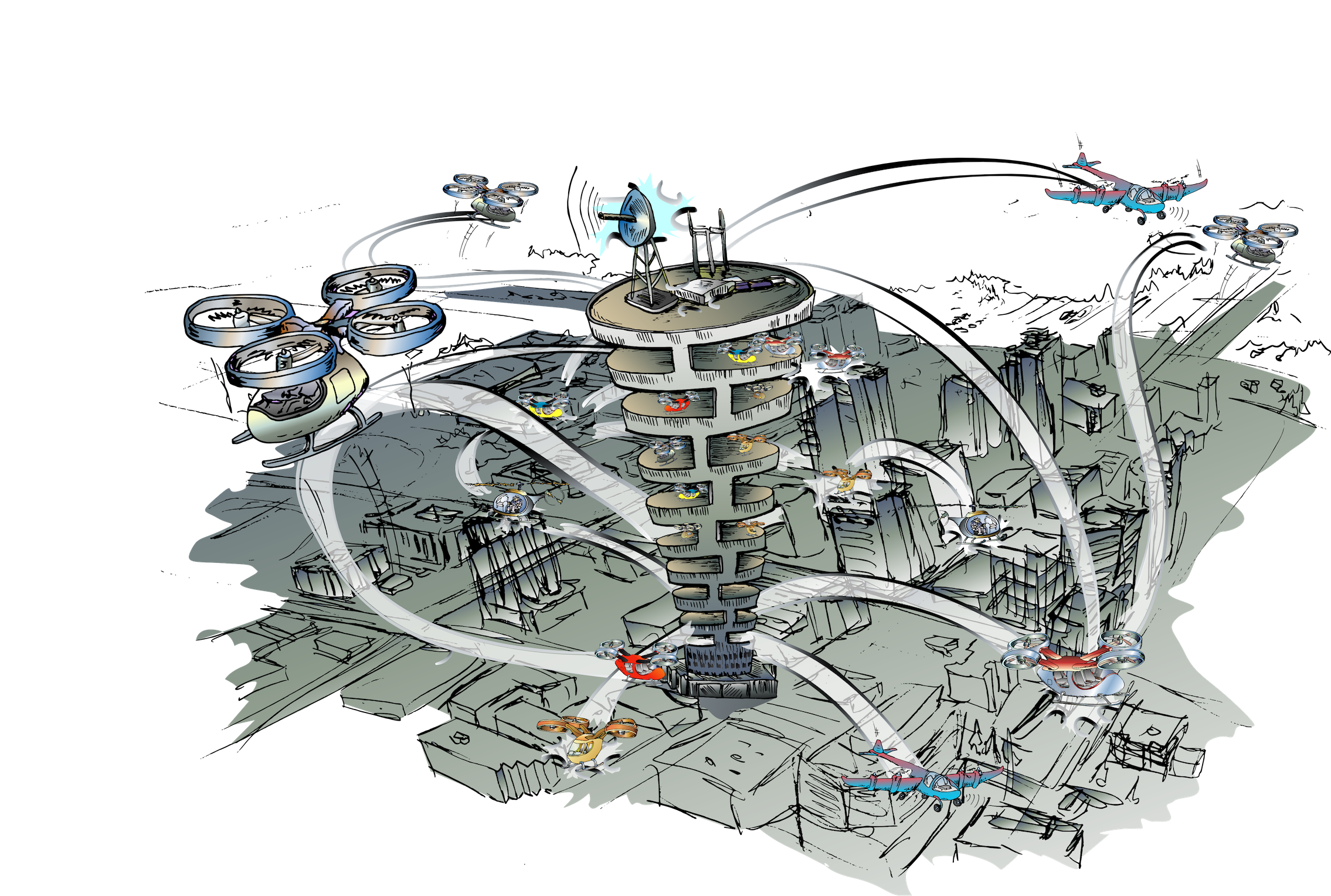}
\caption{The urban airspace of the future will involve multiple platforms operating on various size and timescales. }\label{Fig:UrbanMobility}
\end{center}
\vspace{-.25in}
\end{figure}

Urban air mobility is an emerging transportation framework with the goal of providing on-demand, personal point-to-point transportation within and between the obstacle-rich environments of urban domains by leveraging the airspace above roads and between and above buildings and other physical infrastructure.  Urban air mobility has great promise for alleviating transportation congestion, decreasing travel times, decreasing levels of pollution through the use of next-generation electric aircraft, and producing technology to benefit a wide range of industries; see Fig.~\ref{Fig:UrbanMobility}.  The purpose of this case study is to demonstrate how the component subsystems of urban air mobility have and can benefit from data science.

In order to achieve the promise of point-to-point, on-demand personal transportation, a number of key technological challenges must be addressed.  
These challenges include, but are not limited to:  vehicle design, testing, and certification; vehicle fleet health monitoring; vertical garages (vertiport) design and construction for passenger on- and off-boarding; minor maintenance such as battery charging, fueling, cleaning and major maintenance and repair; logistical scheduling systems, resembling land-based ride-shares, taxis, trains, and buses; air traffic routing, requiring next generation air traffic control; integration of autonomy for guidance, navigation, and control; and autonomous deconfliction, such as automatic dependent surveillance-broadcast (ADS-B).  
Nearly all of these technologies will rely on improved sensor networks, robust data communication and processing, and advanced machine learning algorithms.  
As discussed next, these system elements are at varying levels of maturity and technology readiness.

A large number of companies, both established and startup, have been exploring the development of vehicles suitable for dense and safe personal transportation in urban settings using electric propulsion.  
Traditional fixed wing aircraft require long runways that are spatially infeasible in cities, while helicopter designs are unable to scale to large numbers because of maneuverability and safety issues stemming from the lack of unpowered gliding capability.  
Innovative designs have been proposed over the past decade to achieve the needs of safe, dense operation with high maneuverability, safety, and green fuel.  
The designs of such vehicles require the perspective of multi-disciplinary optimization with regard to aero-servo-elasticity considerations.  
Specifically, there is a tight physical coupling between the unsteady aerodynamics over the flexible airframe, which is further influenced by the propulsion and control surfaces.  
Thus, these components cannot be decoupled and designed independently, making the design space quite rich and challenging. 

Historically, successful innovation of viable airframes has required years of extensive experience.  
The advent of modern computational capabilities will enable faster exploration of the design space through computational fluid dynamics (CFD) and finite element analysis (FEA) tools to simulate vehicle performance in a range of flight and loading conditions. 
These tools, and the design space over which they are leveraged, are inherently data rich. 
While CFD and FEA are effectively already data science tools, the potential exists for more extensive analytics to intelligently reduce the vast design space to a more manageable level.  
An example of how such design tools can impact outcomes can be seen in the multirotor design tool~\cite{du2016computational} that produces unexpected configurations for vehicles with multiple sets of rotating blades like a helicopter.   
While those designs are not likely options for passenger vehicles, the current innovation environment coupled with both the traditional design experience and modern data science tools has led to a rich array of proposed vehicles for urban operation.  
Options being considered include multirotors, short runway vertical take off and landing, and various hybrid rotor/propulsion designs.  
At this time, no single vehicle design is a clear frontrunner, and as with traditional fixed wing aircraft, the most likely outcome will be a series of vehicle designs from companies that will share the market and the urban air space.

As with long-range air transportation, stable operation of fielded fleets of vehicles requires vehicle health monitoring, maintenance systems, and technical support.  
Onboard sensing and data logging capabilities for mechanical systems has been developing rapidly, with many individual subsystems (e.g., power, propulsion, cabin environment) being equipped with data systems that log and either save or regularly transmit system health information to a data server (e.g., Rolls Royce and BMW).  
The sheer volume of data provided by these systems is immense, and appropriate data mining and monitoring tools are needed to flag operation and maintenance issues and to remediate issues that arise during flight.  
Maintenance and service for these vehicles will likely be a bit different than for long-range flight vehicles.  
For aircraft powered by batteries, down time between passengers may require a quick battery charge or replacement.  
For vehicles with combustion engines, fuel tanks will need to be filled.  
Operational inspections will need to take place to ensure passenger safety.   
These tasks must all be performed in such a way as to facilitate efficient vehicle take offs, landings, and ground maneuvering.  
Solutions for efficiently managing these tasks are actively in development for other applications such as autonomous ocean transportation, autonomous ground transportation, and smart city infrastructure.
In addition, vertical \emph{garages} will be key to the successful adoption of urban air transportation, enabling safe takeoff and landing and facilitating efficiencies that might not be possible otherwise.  

Logistical systems appropriate for handling on-demand transportation requests are quite mature.  The advent of ride-share companies, such as Uber and Lyft, have driven the development of effective mobile applications that handle all aspects of the process, from user registry, payment, localization, service connection, tasking, scheduling, routing, and feedback surveys.  
These tools are at a level of maturity that should translate readily to the urban air mobility domain.  
Data science tools have been developed specifically to address these needs and are extensively leveraged for many aspects of the entire logistics system.  
One exception in this technology is air traffic routing.  Current tools for ground transportation rely on a well-structured operational environment which has extensive physical infrastructure (e.g., roads, buildings, bridges, traffic lights, signage, etc) and \emph{rules-of-the-road} that inform and are informed by regulatory authorities.  

A key question regarding urban air transportation is that of whether aircraft will be piloted, unpiloted or a mix of both.  Given the designs under development, the most likely scenario is the hybrid case where a mix of many different types of vehicle piloting operate simultaneously.  Even in the case of human piloted vehicles, a great deal of vehicle autonomy is in operation at any time to manage the myriad subsystems  that are required for a modern flight system.  The dense airspace of the future will likely exacerbate the existing challenges of autonomy.  
Further, the current human-centric air traffic control framework will simply be unable to address the temporal and spatial density needed for urban air mobility to be viable. 
However, autonomous routing, deconfliction, and obstacle avoidance are not yet at a level of maturing to be fielded.  The issues are partly technological and partly policy.  On the technology side, sensors and algorithms are in late stage development (e.g., ADS-B), but they have not yet been adopted into standard practice.  On the policy side, work has been underway for Next Generation Air Traffic Control (NextGen ATC) for the past two decades, but final adoption has not yet reached final stages that were planned for 2025.  To realize the NextGen ATC system, data science techniques are needed to provide real-time routing of all vehicles in the same way that ground-based transportation systems utilize.

\section{Outlook}\label{Sec:Outlook}
This paper has provided a high-level review and roadmap for the uses of data science and machine learning in the aerospace industry.  
Aerospace engineering is a data-rich endeavor, and it is poised to capitalize on the big-data and machine learning revolution that is redefining the modern scientific, technological, and industrial landscapes.  
We have outlined a number of high-priority opportunities in aerospace design, manufacturing, testing, and services that all may be enabled by advanced data science algorithms.  
For example, future data-driven algorithms may enable the design of new super-materials, streamlined flight testing for safer and more flexible designs, and enhanced manufacturing processes including standardization, predictive assembly, and non-destructive inspection.  
The array of potential applications is both riveting and at times may be overwhelming. 
Some of these advanced technologies, such as predictive shimming and assembly, are already experiencing considerable success in commercial production.  
Other applications will be enabled by the digital twin of the future, with advanced optimization and learning algorithms seamlessly fusing multi-physics and multi-fidelity models with real-world data streams. 
These models will continuously improve, with potentially radical implications for the entire aerospace production pipeline.  

There are clear parallels between the recent rise of data science and the transformative impact of scientific computing beginning in the 1960s.  
Computational science began as a specialized focus and has since evolved into a core engineering competency across many disciplines.  
Likewise, data science and machine learning proficiency will be expected as a core competency in the future workforce, highlighting the need for robust education initiatives focused on data-driven science and engineering~\cite{Brunton2019book}\footnote{See databookuw.com for video lectures, syllabi, and code that are tailored for a science and engineering audience.}. 
Data-driven aerospace engineering will also require changes to how teams of researchers and engineers are formed and how decisions are made.  
A recent Harvard Business Review article~\cite{patil2012data} pointed out the need for data scientists to be integrated into the decision making process, likening this to having Spock on the bridge in Star Trek.  
The last challenge to discuss is how to develop a \emph{data-first} culture, which is imperative to fully harness the potential of data-enabled engineering. 
A data-first culture requires enterprise-wide education and adoption of principles that promote cooperation and data sharing, reproducible practices, common terminology, and an understanding of the value of data and the promise and limitations of machine learning algorithms.  
Moreover, this transformation will require ongoing curiosity and deep thinking from engineers and leaders about how processes might be improved with emerging technologies.  

Current and future aircraft programs will be increasingly enabled by a wealth of data. 
There is low-hanging fruit in current aircraft programs that may provide immediate benefits with existing data.  
However, the full potential of data-enabled aerospace engineering will take generations of aircraft programs to realize.  
New programs under development, such as the new midsize airplane (NMA), will provide an opportunity to develop and test entirely new design, manufacturing, and testing capabilities.  
The digital twin will improve design and testing cycles through a more wholistic data-driven model of critical processes and their interactions.  
These integrated models will rely on the integration of myriad data sources into a digital thread.  
Lessons learned may then be leveraged into more seamless integration of data into the aerospace industry for future programs, such as the future small aircraft (FSA). 
Much as information technology companies like Google are valued based on their data, aerospace giants will learn to extract value and competitive advantage by leveraging their wealth of data. 

Never before has there been an architecture problem of this size, requiring the integrated efforts of information technology experts, aerospace domain engineers, and data science teams.  
The aerospace industry currently generates tremendous volumes of data throughout a product life cycle, but the data storage systems are not always designed to have their data extracted, much less at near real-time rates.  
Data must be seamless to access while maintaining security and control; software must enable analytics while performing its primary objective.  
Data must be integrated while being served by decades-old systems.  
Data exploration must be near-effortless and allow for analysis of complex systems.  
Analytic results must be fed back into the systems providing suggestions or predictions must be resolved without forcing all data to be migrated or new systems wholesale be replaced. 
There is also the risk of a data mortgage and paralysis, where more effort is spent in collecting and curating data than analyzing it.  
This risk motivates a shift from \emph{big data} to \emph{smart data}, where edge-computing and sparse/robust algorithms are used to extract key data features in real time.  
Moving from a process largely grounded in storing data in local silos to openly sharing data will be difficult and will motivate entirely new incentive structures for engineers.
Teams must come to identify the value they provide in part based upon the data they generate and curate.  

It is also important to reiterate the need to develop machine learning algorithms that are specifically tailored for the aerospace industry.  
Machine learning must be demystified:  it is not a magic wand, but rather a growing corpus of applied optimization algorithms to build models from data.  
These models are only as good as the data used to train them, and great care must be taken to understand how and when these models are valid. 
Most machine learning algorithms are fundamentally interpolative, and extrapolation algorithms are both rare and challenging.  
Because of the need for reliable and certifiable algorithms, it is critical that physics is baked into machine learning algorithms.  
Many of the concepts related to data analytics detailed within this paper constitute novel applications that may not support a direct showing of compliance to existing regulations. 
Regulatory compliance is a critical challenge for emerging approaches based on data analytics. 
Historically, civil aviation requirements have been created and revised in response to negative events in the industry: primarily in reaction to accidents and incidents. 
Further, these requirements are often made under the expectation that they are deterministic in nature. 
As non-deterministic systems become introduced into commercial aviation systems, the basic regulatory philosophy will not support the current methods of compliance. 
Anywhere that data analytics are leveraged in the certification process, the same philosophical issues will need to be addressed. In place of a deterministic answer, statistical approaches, and potentially models with dis-similar architectures, or other approaches not yet identified will become a necessity. Fortunately, statistics are used in various areas of aircraft certification, related to aircraft safety assurance following single and multiple failure events.

Powered human flight is one of the greatest achievements in the history of humankind, realizing the culmination of thousands of years of science fiction, and having a profound impact on the past century.  
The next century of aerospace engineering will challenge us to envision and realize a new science fiction future based on the breathtaking array of new data-enhanced technologies.   

 \section*{Acknowledgments}
 We would like to acknowledge several individuals and organizations within the University of Washington and The Boeing Company:  The Boeing Advanced Research Center (BARC) and its directors Per Reinhall, Santosh Devasia, and Sam Pedigo; Tia Benson Tolle, Steve Chisholm, Chris Schuller, and Todd Zarfos at Boeing; Ashis Banerjee, Bing Brunton, Ramulu Mamidala, and Navid Zobeiry at UW; the eScience Institute and its directors Ed Lazowska and Magda Balazinska.
We would also like to give special thanks to Mike Bragg for his support as Dean of the College of Engineering at UW and Greg Hyslop, Lynne Hopper, and Nancy Pendleton at Boeing for expanding the scope of this collaboration.  We are also grateful for advice and interviews with Ryan Smith, Laura Garnett, Phil Crothers, Nia Jetter, Howard McKenzie, and Darren Macer.  Finally, we are indebted to the creative efforts of Michael Erickson and the envisioneering team who translated our engineering discussions into many of the graphics in this paper. 

 \begin{spacing}{.9}
\footnotesize{
 \setlength{\bibsep}{1.5pt}

 }
 \end{spacing}
\end{document}